\documentclass[10pt,twocolumn,letterpaper]{article}

\PassOptionsToPackage{table}{xcolor}

\usepackage[pagenumbers]{cvpr}

\usepackage{amsfonts}
\usepackage{algorithmic}
\usepackage{textcomp}
\usepackage{bm}
\usepackage{multirow}
\usepackage{threeparttable}
\usepackage{makecell}
\usepackage{placeins}
\usepackage{float}
\usepackage{microtype}

\newcommand{\safeincludegraphics}[2][]{%
  \IfFileExists{#2}{%
    \includegraphics[#1]{#2}%
  }{%
    \fbox{%
      \parbox[c][1.55in][c]{0.94\linewidth}{%
        \centering\small Missing figure\\[2pt]
        \texttt{\detokenize{#2}}%
      }%
    }%
  }%
}

\newcommand{\ra}[1]{\renewcommand{\arraystretch}{#1}}

\def\BibTeX{{\rm B\kern-.05em{\sc i\kern-.025em b}\kern-.08em
    T\kern-.1667em\lower.7ex\hbox{E}\kern-.125emX}}

\definecolor{cvprblue}{rgb}{0.21,0.49,0.74}
\usepackage[breaklinks,colorlinks,allcolors=cvprblue]{hyperref}

\def\confName{CVPR}

\title{TAME: Temporal-Aware Mixture-of-Experts for Text-Video Retrieval}

\author{
Uicheol Jung$^{1}$ \quad
Juyoung Hong$^{1}$ \quad
Hojung Kwon$^{2}$ \quad
Yukyung Choi$^{1}$\\[0.4em]
$^{1}$Sejong University, Seoul, Republic of Korea
\qquad
$^{2}$Wisenut, Seongnam, Republic of Korea\\[0.3em]
{\tt\small ucjung@rcv.sejong.ac.kr \quad ykchoi@rcv.sejong.ac.kr}
}

\begin{document}
\maketitle

\begin{abstract}
Text–Video Retrieval (TVR) retrieves videos that match a natural-language query, but extending image–text models such as CLIP to videos is fundamentally limited by the lack of temporal modeling. Videos exhibit frame-wise heterogeneity in appearance and motion, and compressing all frames into a single representation often obscures temporal structure and semantic transitions. To address this, we propose \textbf{Temporal-Aware Mixture-of-Experts for Text-Video Retrieval (TAME)}, a CLIP-based framework that jointly models frame-level structure and temporal relations. First, we integrate sparse Mixture-of-Experts (MoE) layers into both CLIP encoders and apply frame-consistent routing on the vision branch so that experts specialize according to frame-level visual patterns while preserving the original vision--language alignment. Second, we introduce Frame–Temporal (FT) tokens that aggregate global cross-frame information and feed it back to each frame, enabling the visual encoder to capture long-range temporal dependencies without harming local details. Third, we design a Cross-Temporal Interaction and Aggregation (CTIA) module that refines frame-wise sentence–video similarities through staged temporal filtering and fusion. Experiments on standard TVR benchmarks show that TAME consistently improves over CLIP-based baselines. On MSR-VTT, it improves R@1 by 4.0 over CLIP4Clip, and also achieves consistent gains on DiDeMo, MSVD, LSMDC, and ActivityNet. The code is available at \url{https://github.com/sejong-rcv/TAME}.
\end{abstract}
\section{Introduction} \label{sec:introduction}

With the explosive growth of video content across social platforms, entertainment services, and multimedia archives, the ability to retrieve relevant videos from natural-language queries, known as Text-Video Retrieval (TVR), has become increasingly critical. Unlike traditional metadata-based search, TVR demands fine-grained multimodal understanding, where both visual signals and linguistic semantics must be aligned at the appropriate spatiotemporal granularity. This capability is essential for applications such as content recommendation, video understanding, and large-scale retrieval systems deployed in industrial settings.

\begin{figure*}[!t]
  \centering
  \safeincludegraphics[width=0.8\textwidth]{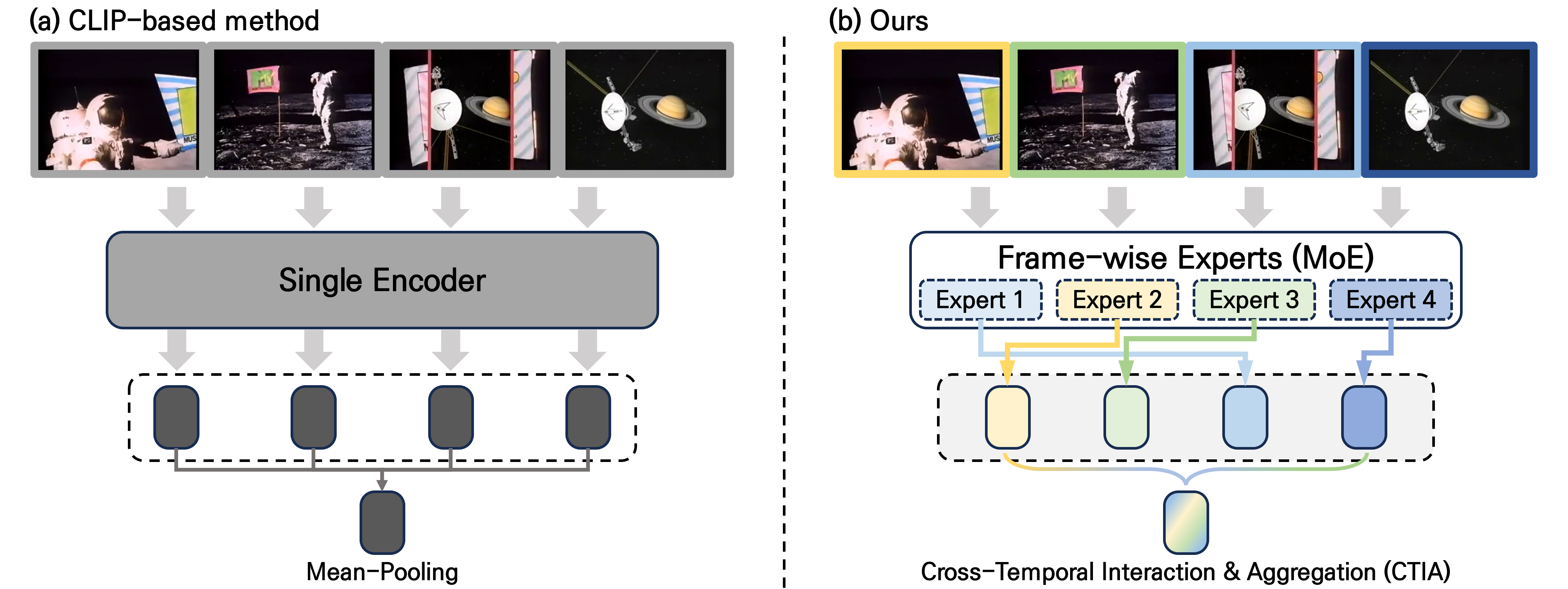} 
  \caption{Comparison between previous method and ours. (a) Previous methods encode all frames with a single encoder and average them uniformly, which overlooks temporal dependency and frame importance. (b) Our TAME performs frame-wise expert routing to model visual diversity and employs Cross-Temporal Interaction and Aggregation (CTIA) to adaptively weight and integrate frame representations according to their temporal context, yielding a robust video-level feature.}
  \vspace{-3mm}
  \label{fig:introduction}
\end{figure*}

Recent TVR research has been accelerated by CLIP-style dual encoders~\cite{CLIP}, which leverage strong image–text alignment learned from large-scale contrastive pretraining. Previous studies~\cite{CLIP4Clip, XCLIP, XPool} demonstrate that transferring image–text representations to the video domain provides a powerful foundation for retrieval. However, as illustrated in Fig.~\ref{fig:introduction}(a), conventional CLIP-based approaches aggregate all frame embeddings into a single video representation, relying primarily on mean pooling. This strategy overlooks temporal structure, compresses heterogeneous frame-level patterns, and provides no mechanism to adaptively weight frames based on semantic relevance. Consequently, CLIP-based models struggle to fully capture the temporal progression and semantic variability that characterize videos.

Recent and representative studies~\cite{TS2Net, CLIPViP, CenterCLIP, TeachCLIP, MVAdapter} attempt to mitigate this issue by introducing temporal aggregation/attention, cross-frame token interaction, or token selection/clustering modules. Yet, these approaches still rely on a single dense encoder pathway, forcing all frames through the same feature transformation. This architectural bottleneck limits representational capacity and prevents the model from adapting to heterogeneous visual patterns—such as static versus dynamic scenes, background shifts, or object composition changes across frames. The lack of specialization becomes more pronounced as video content becomes more visually diverse.

To address these challenges, we propose Temporal-Aware Mixture-of-Experts for Text-Video Retrieval (TAME), a unified framework designed to capture frame-level variability and improve temporal aggregation for retrieval. As illustrated in Fig.~\ref{fig:introduction}(b), TAME replaces dense FFNs in both the text and vision encoders with sparse Mixture-of-Experts (MoE) layers, enabling frame-consistent expert routing that expands representational diversity while preserving CLIP alignment. Importantly, extending MoE to the text encoder supports semantic specialization on the language side, allowing the model to capture diverse linguistic patterns and maintain balanced cross-modal capacity. In addition, we introduce Frame–Temporal (FT) tokens that propagate compact temporal context across frames while keeping patch-level spatial structure intact, and we design a Cross-Temporal Interaction and Aggregation (CTIA) module that refines frame–sentence similarity into a coherent video-level score.

Our key contributions are summarized as follows:
\begin{enumerate}
    \item \textbf{Sparse Mixture-of-Experts architecture tailored for TVR.} We introduce a sparse MoE framework specifically tailored for Text–Video Retrieval. By integrating MoE layers into both the text and visual encoders and adopting frame-consistent routing, the model captures heterogeneous frame-level visual patterns that are difficult to model using a single shared encoder.
    \item \textbf{Frame–Temporal tokens for lightweight temporal propagation.} We propose FT tokens that encode compact temporal context and interact with frame and patch embeddings, enabling the model to share temporal information across frames without modifying spatial token structure.
    \item \textbf{Sentence-conditioned cross-temporal aggregation.} To address temporal inconsistency introduced by frame-wise expert routing, we propose a Cross-Temporal Interaction and Aggregation (CTIA) module that refines frame–sentence similarity into a coherent video-level representation.
    \item \textbf{Consistent performance gains across TVR benchmarks.} Evaluation on MSR-VTT, DiDeMo, MSVD, LSMDC, and ActivityNet shows that the proposed architecture consistently improves over CLIP-based baselines.
\end{enumerate}

Ultimately, TAME retains CLIP’s strong vision–language alignment while resolving its inherent temporal limitations, achieving more precise and resilient text–video matching.

\begin{figure*}[!t]
  \centering
  \safeincludegraphics[width=0.7\textwidth]{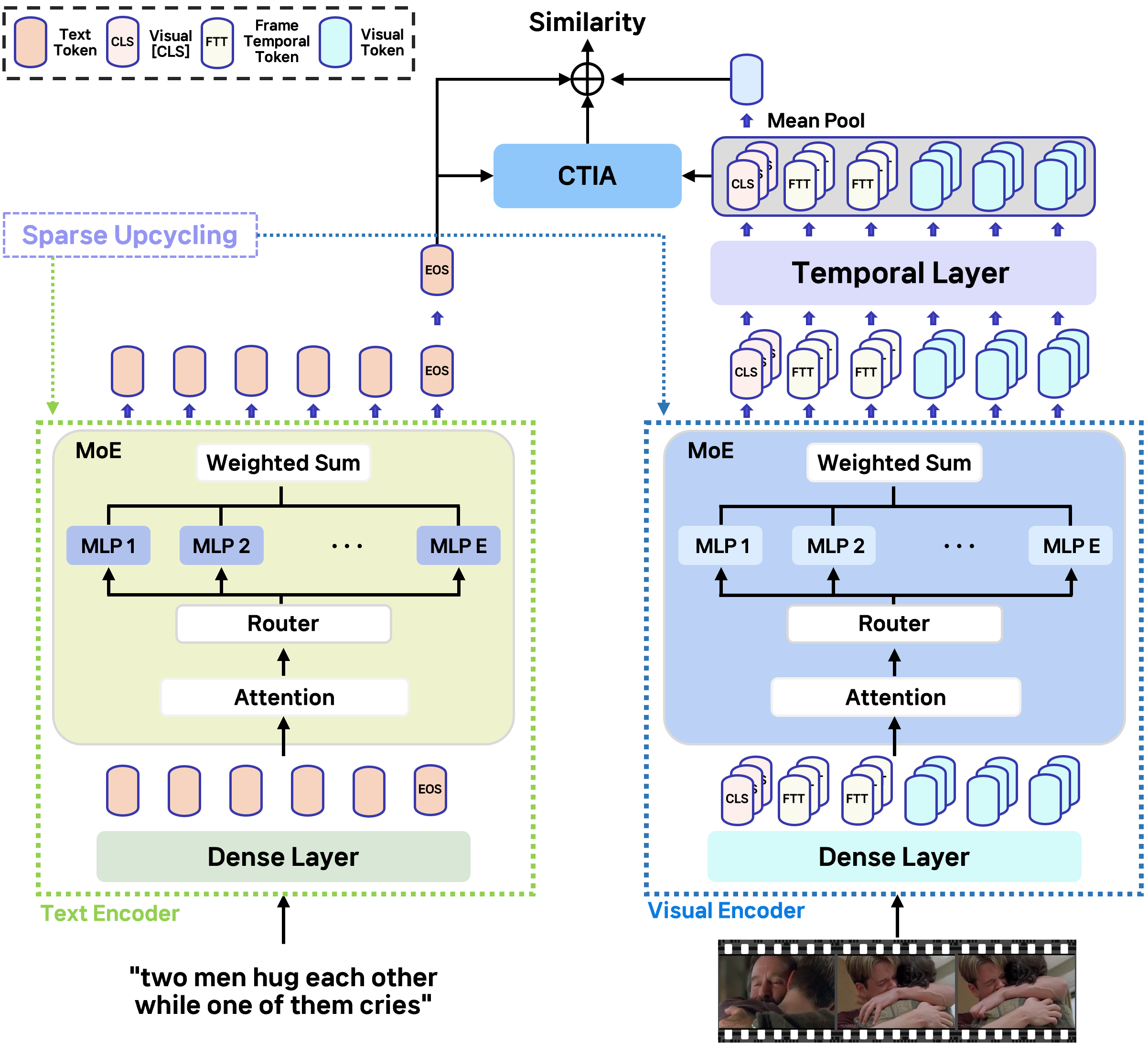} % 0.88로 축소
  \caption{
  TAME overview. Based on CLIP dual encoders, we introduce three add-on modules to enhance text--video retrieval. \textbf{(i) MoE-Enhanced CLIP:} we apply sparse upcycling that replaces selected FFN blocks with MoE layers to expand encoding capacity; experts are initialized from the dense checkpoint, while routers are randomly initialized. \textbf{(ii) Frame--Temporal (FT) tokens:} videos are sampled into frames and encoded into frame-level representations, which are further refined by injecting lightweight cross-frame context in the visual branch. \textbf{(iii) CTIA:} the CTIA module performs sentence-conditioned frame alignment and aggregates frame evidence to estimate the final frame-aware similarity. In addition to the frame-aware branch, frame representations are mean-pooled to form a video representation for global alignment with the sentence embedding.}
  \label{fig:method}
\end{figure*}

% Preamble:
% \usepackage{xcolor}

\section{Related Work}

\subsection{CLIP-based Text--Video Dual Encoders}
Recent text--video retrieval (TVR) pipelines commonly adopt CLIP-style dual encoders that map text and video into a shared embedding space and compute similarities for ranking.
Representative approaches such as CLIP4Clip~\cite{CLIP4Clip}, CLIP2Video~\cite{CLIP2Video}, CenterCLIP~\cite{CenterCLIP}, 
X-CLIP~\cite{XCLIP},
TS2-Net~\cite{TS2Net}, X-Pool~\cite{XPool}, CLIP-ViP~\cite{CLIPViP}, Prompt Switch~\cite{PromptSwitch}, Cap4Video~\cite{Cap4Video}, TeachCLIP~\cite{TeachCLIP} differ in how they pool frame features, parameterize sentence--video alignment, and incorporate lightweight temporal modules, while sharing a pretrained CLIP~\cite{CLIP} backbone and a similarity head over sentence and frame-level features.
However, a recurring limitation in this line is that video-level decisions are still often derived from relatively shallow temporal aggregation (e.g., pooling or local temporal blocks), which weakens cross-frame interaction and makes the final score vulnerable to temporally noisy frames that are irrelevant to the query.
This issue becomes more pronounced when the model must decide \emph{which frames constitute valid evidence} for the sentence, since naive pooling can over-count spurious matches and under-utilize temporally consistent cues.
In addition, while improving representation capacity could alleviate such brittleness, dense scaling of the encoders is expensive and does not directly guarantee better \emph{video-level} evidence consolidation.
Motivated by these gaps, we introduce a sparse Mixture-of-Experts (MoE) into the CLIP encoders to expand representational capacity with low active compute via sparse gating, and we further propose a Cross-Temporal Interaction \& Aggregation (CTIA) module that explicitly consolidates per-frame sentence--frame scores into a single video-level logit by propagating evidence across neighboring frames and encouraging temporal consistency, thereby making the retrieval decision less sensitive to frame-level noise.

\subsection{Mixture-of-Experts in Vision and Language}
Deep learning has shown that scaling model, data, and compute improves within-distribution generalization. Early studies consistently reported reductions in test-in-distribution loss as scale increases~\cite{deepscaling}. In parallel, large language and multimodal models~\cite{GPT3},~\cite{CLIP},~\cite{Flamingo} showed broad transfer and prompt-based reasoning, and ``scaling-law'' studies formalized power-law links between model/data/compute and test loss~\cite{Scalinglaws},~\cite{Trainingcomputeoptimal}.
However, for many large \emph{dense} models in practice, memory/communication/power costs grow steeply
with scale, often reducing training efficiency. To mitigate these limitations, sparsely activated
Mixture-of-Experts (MoE) architectures~\cite{sparseMoE} have gained attention: a router
selects a small subset of experts per input, thereby expanding total parameter capacity while
\emph{aiming} to keep per-token active compute roughly constant under typical capacity settings.

In vision, V-MoE~\cite{VMoE} extends sparse MoE to Vision Transformers by replacing selected FFN blocks with expert pools and routing \emph{patch tokens} to a small number of experts, enabling substantial capacity scaling with nearly constant active computation.
For videos, recent works explore applying conditional routing to spatiotemporal transformers (e.g., ViViT-style backbones) to dynamically allocate computation across visual tokens under a compute budget~\cite{mone}.

Recent attempts to apply MoE to CLIP further support this direction: CLIP-MoE~\cite{CLIPMoE} uses
multiplet upcycling to lightly fine-tune complementary CLIP branches as experts and combine them via
MoE, showing consistent gains across several public zero-shot classification/retrieval benchmarks and
as an MLLM backbone without substantially increasing compute; CLIP-UP~\cite{CLIP-UP} provides a simple,
efficient recipe to upcycle a pretrained dense CLIP into a sparse MoE, reusing existing weights to
achieve a stronger performance--cost balance under modest additional training.

Different from standard vision/video MoEs that primarily target recognition accuracy--efficiency trade-offs, our setting is text--video retrieval with a CLIP-style dual-encoder, where routing should preserve cross-modal alignment. Accordingly, TAME designs routing signals tailored to retrieval (text-token routing in the text branch and frame-wise routing in the visual branch) while maintaining the pretrained vision--language alignment.

\subsection{Hashing-based Content-Based Medical Image Retrieval}
Beyond CLIP-style TVR, a parallel retrieval literature studies compact representations for efficient indexing, particularly hash-code based content-based image retrieval (CBIR).
In medical imaging, a number of methods learn discriminative deep features and generate hash codes to enable fast search in large-scale archives, including deep feature selection guided Siamese hashing for medical image retrieval~\cite{Hashcode},
class-driven medical image retrieval using hash codes of deep features~\cite{Classdriven},
and attention-based end-to-end CNN frameworks for X-ray image retrieval~\cite{CNNXray}.
These works highlight a complementary set of challenges and solutions: they typically focus on single-modality image-to-image retrieval and emphasize compact/binary hashing for storage and efficient indexing, but they do not address cross-modal semantic alignment (text--vision) nor the temporal evidence aggregation that is intrinsic to videos.
Consequently, although hashing-based medical CBIR methods are relevant as retrieval-oriented representation learning, their core gaps relative to our setting include limited support for cross-modal grounding and the absence of mechanisms for consolidating temporally distributed evidence.
In contrast, our work targets cross-modal text--video retrieval and addresses the video-specific decision bottleneck by combining capacity-efficient scaling (sparse MoE in CLIP encoders) with temporally consistent evidence aggregation (CTIA), which directly improves video-level ranking robustness rather than optimizing binary codes for single-image indexing.

\section{Method}

In this section, we introduce the overall methodology of the proposed framework TAME. The entire architecture is illustrated in Fig.~\ref{fig:method}. 
First, Sec.~\ref{Sec:CLIPEncoder} describes CLIP-based dual encoders that extract aligned text and frame-level video representations.
Then, Sec.~\ref{Sec:3B} introduces MoE-Enhanced CLIP, which replaces selected dense FFNs with sparse experts to enhance representation capacity.
Next, Sec.~\ref{Sec:3C} presents Frame--Temporal (FT) tokens, a lightweight temporal operator that injects cross-frame context into the visual representations before computing framewise similarities.
Sec.~\ref{Sec:CTIA} then introduces CTIA, which performs sentence-conditioned aggregation over frames to complement the baseline similarity.
Finally, Sec.~\ref{Sec:3E} specifies the training objective and joint optimization of the contrastive and auxiliary losses.

\noindent\textbf{Formulation overview and linkage to our contributions.}
We consider text--video retrieval with a text query $T$ and a video $V=\{I_n\}_{n=1}^{F}$ consisting of $F$ sampled frames.
The CLIP text encoder outputs token embeddings $\mathbf{e}\in\mathbb{R}^{(L_t+1)\times C}$, and we use the $[\mathrm{EOS}]$ embedding as the global sentence representation, denoted by $q=e_{[\mathrm{EOS}]}\in\mathbb{R}^{C}$.
The visual encoder maps each frame $I_n$ to a frame-level representation $f_n\in\mathbb{R}^{C}$ (the per-frame $[\mathrm{CLS}]$ embedding), yielding the frame sequence $\mathbf{F}=\{f_n\}_{n=1}^{F}\in\mathbb{R}^{F\times C}$.
We compute framewise sentence--frame evidence $\mathbf{s}\in\mathbb{R}^{F}$ by $s_n=\cos(q,f_n)$ and define $\mathbf{s}=(s_1,\ldots,s_F)$.
This framewise evidence quantifies which frames are directly relevant to the query and serves as the input signal for CTIA to model cross-frame interactions and perform decision-level temporal evidence aggregation.

Prior CLIP4Clip-style baselines primarily make the retrieval decision using a video-level similarity computed from an aggregated video embedding.
In our framework, we additionally exploit the framewise evidence $\mathbf{s}$ and define the final retrieval logit as a function of both the sentence and the frame sequence, i.e., $S_{\text{final}}=\Phi(q,\mathbf{F})$.
We instantiate $\Phi(\cdot)$ via CTIA by defining a sentence-conditioned aggregation function $\mathrm{Agg}(\mathbf{s},\mathbf{F})$ and combining its output with the baseline similarity to obtain $S_{\text{final}}$ (see Sec.~\ref{Sec:CTIA} for the explicit formulation).

This formulation clarifies how each proposed component contributes.
MoE-Enhanced CLIP modifies the encoders that produce $q$ and $\mathbf{F}$ by replacing selected dense FFNs with sparse experts under top-$K$ routing (Sec.~\ref{Sec:3B}).
Frame--Temporal (FT) tokens refine $\mathbf{F}$ by injecting cross-frame context in the visual branch prior to computing $\mathbf{s}$ (Sec.~\ref{Sec:3C}).
CTIA defines $\mathrm{Agg}(\cdot)$ to transform $\mathbf{s}$ into frame weights and a decision-level score used in $S_{\text{final}}$ (Sec.~\ref{Sec:CTIA}).
Finally, we use $S_{\text{final}}$ as the similarity logit to construct the batch-wise matrix $S_{ij}=S_{\text{final}}(v_i,q_j)$ and optimize the retrieval objective (Sec.~\ref{Sec:3E}).

\subsection{CLIP Encoder} \label{Sec:CLIPEncoder}
Following prior work~\cite{CLIP4Clip}, we adopt a dual-encoder architecture based on the pre-trained CLIP~\cite{CLIP} model to extract aligned text–video representations within a shared embedding space.

\textbf{Text Encoder.} Given a text query $T$, the CLIP text encoder processes this sequence and outputs a sequence of token embeddings $\textbf{e} = \{e_1, e_2, ..., e_{L_t}, e_{[\text{EOS}]}\} \in \mathbb{R}^{(L_{t}+1) \times C}$, where $L_t$ denotes the number of textual tokens in $T$, and $C$ is the feature dimension of each token embedding. Here, $e_{[\text{EOS}]} \in \mathbb{R}^C$ corresponds to the embedding of the special $[\text{EOS}]$ token, which is used as the global sentence representation.

\textbf{Visual Encoder.} Given a video $V$ consisting of $F$ sampled frames, each frame is processed by the CLIP visual encoder. For the $n$-th frame, we partition it into $M$ patches and prepend a [CLS] token, resulting in a set of patch embeddings $P_n = \{p_n^1, p_n^2, \dots, p_n^M\} \in \mathbb{R}^{M \times C}$, where all visual tokens share the same feature dimension $C$ as textual tokens. The embedding of the [CLS] token serves as the frame-level representation $f_n \in \mathbb{R}^C$. Aggregating all frame embeddings yields the video-level sequence $\mathbf{F}=\{f_1, f_2, \dots, f_F\} \in \mathbb{R}^{F \times C}$.

Both textual and visual embeddings are further mapped into a shared embedding space, allowing direct similarity computation between the global sentence embedding $e_{[\text{EOS}]}$ and the video-level sequence $\mathbf{F}$ for downstream similarity computation. This CLIP-based feature extraction serves as the backbone for our proposed MoE-Enhanced CLIP, which augments the encoder layers with MoE modules to better capture diverse frame-level semantics and temporal dependencies.
    
\begin{figure}[!t]
  \raggedright 
  \safeincludegraphics[width=1\linewidth]{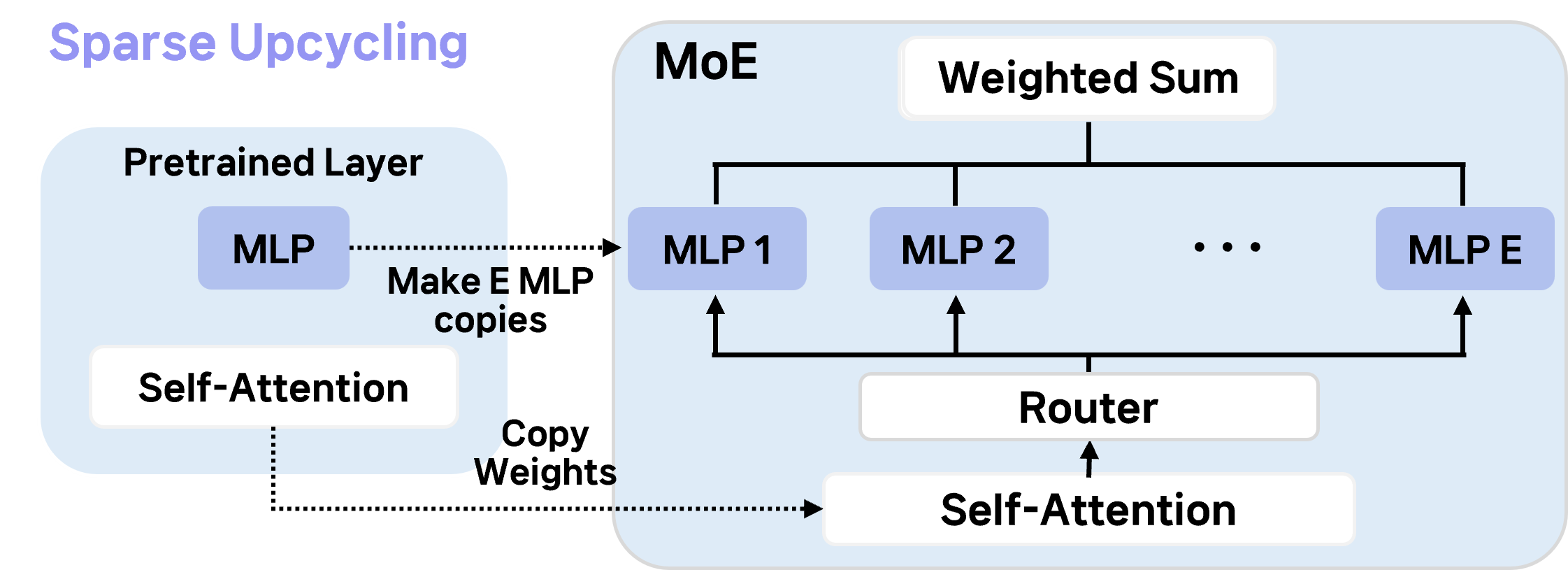}
  \caption{{\textbf{Sparse upcycling (MoE-Enhanced CLIP).} We replace the pretrained image-layer FFN (MLP) with a sparse MoE layer composed of \(E\) cloned MLP experts. The expert weights are copied from the dense checkpoint to preserve CLIP initialization, while the router is randomly initialized to enable expert specialization during training.}}

  \label{fig:fig3}
\end{figure}

\subsection{MoE-Enhanced CLIP} \label{Sec:3B}

In Text-Video Retrieval, the encoders must handle highly diverse visual and linguistic patterns. Accordingly, it is important to enhance CLIP’s encoding capacity while preserving its pretrained vision–language alignment. Sparse Mixture-of-Experts (MoE) offers a principled way to achieve this~\cite{VMoE, SwitchTransformer, CLIP-UP, CLIPMoE}: it replaces the dense feed-forward network (FFN) with a set of parallel MLP experts and routes each token to only a small subset of them, thereby expanding representational capacity without increasing computation.

To this end, we follow prior MoE-augmented Transformer studies~\cite{CLIP-UP, CLIPMoE} and integrate sparse MoE layers into both CLIP encoders. Among the $L$ Transformer layers, we replace a subset of the FFNs with MoE layers, each consisting of $N_E$ position-wise MLP experts $\{E_i\}_{i=1}^{N_E}$ that take the role of the original FFN.

A lightweight router $G(\cdot)$ assigns each input token a score for every expert. We select only the top-$K$ experts to enforce sparsity, normalize the selected scores with a softmax, and obtain the MoE output as a weighted sum of the selected expert outputs. Following~\cite{CLIP-UP, CLIPMoE}, each expert is initialized by copying the pretrained CLIP FFN, while the router is randomly initialized. This initialization maintains compatibility with the pretrained checkpoint and allows experts to gradually specialize during training. The overall MoE block architecture is illustrated in Fig.~\ref{fig:fig3}.

\textbf{Text Branch.} Text-side routing is performed \emph{per token}. For a given MoE layer, let $h_j \in \mathbb{R}^{C}$ denote the representation of the $j$-th token. The text router $G_t(\cdot)$ produces expert logits for $h_j$, from which only the top-$K$ entries are retained and normalized. We define the resulting sparse gating vector as: 
\begin{equation} 
\theta_j = \mathrm{softmax}\!\left(\operatorname{TopK}\!\left(G_t(h_j)\right)\right). 
\end{equation} 
Here, $\theta_{j,i}$ denotes the routing weight assigned to expert $E_i$ for token $j$. The MoE output is then computed as a gated mixture over the experts: \begin{equation} 
    y_j = \mathrm{MoE}(h_j) = \sum_{i=1}^{N_E} \theta_{j,i} \cdot E_i(h_j). 
\end{equation}

\textbf{Visual Branch.} Unlike text, video inputs contain dense spatial patches within each frame, and vanilla MoE routing assigns experts independently to every patch. Such patch-level routing often results in inconsistent expert selection within the same frame, failing to capture coherent frame-level semantics. To mitigate this, we introduce \emph{frame-wise} routing. For frame $n$, let $\tilde{f}_n \in \mathbb{R}^{C}$ denote the frame-level [CLS] token, and let $X_n \in \mathbb{R}^{M \times C}$ denote its $M$ patch tokens. The video router $G_v(\cdot)$ produces expert logits from $\tilde{f}_n$, and the resulting sparse gating vector is defined as: 
\begin{equation} 
    \theta_n = \mathrm{softmax}\!\left(\operatorname{TopK}\!\left(G_v(\tilde{f}_n)\right)\right).
\end{equation} 
The same gating weights are then applied uniformly to all patch tokens, and the MoE output for frame $n$ is computed as: 
\begin{equation} 
    y_n = \mathrm{MoE}(X_n) = \sum_{i=1}^{N_E} \theta_{n,i} \cdot E_i(X_n). 
\end{equation} 
This frame-consistent routing enforces intra-frame coherence, suppresses patch-level routing noise, and enables experts to specialize in distinct frame-level visual patterns, which is crucial for modeling temporal structure.

\textbf{Auxiliary Loss.} While the routing schemes above determine expert selection, they do not enforce balanced expert utilization. Following the Switch Transformer~\cite{SwitchTransformer}, we adopt an auxiliary load-balancing loss to prevent routing collapse. For an MoE layer, let $z$ denote an input token and $q_i(z)$ the router probability for expert $E_i$. The loss $\mathcal{L}_{\text{aux}}$ is defined as: 
\begin{equation} 
    \mathcal{L}_{\text{aux}} = \alpha \cdot N_E \sum_{i=1}^{N_E} \rho_i \, \pi_i, 
\end{equation} 
where the hard routing frequency $\rho_i$ and the mean router probability $\pi_i$ are given by: 
\begin{equation} 
    \rho_i 
    = \frac{1}{|B|}\sum_{z \in B} \mathbf{1}\!\left\{\operatorname*{arg\,max}_j q_j(z) = i\right\}, \end{equation} \begin{equation} \pi_i = \frac{1}{|B|}\sum_{z \in B} q_i(z). 
\end{equation} 
Here, $B$ denotes the set of routed tokens in a batch. Following prior work~\cite{SwitchTransformer}, we set $\alpha = 0.01$. Since both the text and video encoders contain MoE layers—with token-wise routing for text and frame-wise routing for video—we compute the auxiliary load-balancing loss separately for each encoder. We denote these losses as $\mathcal{L}_{\text{aux}}^T$ and $\mathcal{L}_{\text{aux}}^V$, respectively.

\subsection{Frame-Temporal Tokens for Vision MoE} \label{Sec:3C}
CLIP’s vision encoder processes each frame independently, which prevents it from capturing \emph{temporal continuity} (e.g., actions unfolding over time) and \emph{cross-frame dependencies} such as object re-appearance or scene transitions. This limitation is problematic for video understanding, where global temporal context is essential. To address this, we introduce a set of Frame–Temporal (FT) tokens, inspired by the video proxy in CLIP-VIP~\cite{CLIPViP}. For each frame $n$, we insert $H$ learnable FT tokens $t_n = \{t_n^1, \dots, t_n^H\}$ between the frame-level $[\mathrm{CLS}]$ token $f_n$ and the patch tokens $P_n$. These FT tokens serve as compact carriers of inter-frame context and later accumulate global information through both local and global attention mechanisms. The initial input sequence for frame $n$ is then formed as:
\begin{equation}
    \mathbf{T}_n^{(0)} = \operatorname{concat}\!\big(\,f_n,\; t_n,\; P_n \big),
\end{equation}
where $\mathbf{T}_n^{(0)}$ contains one frame-level CLS token, $H$ FT tokens, and $M$ patch tokens.

Non-MoE blocks follow the standard CLIP Transformer, refining intra-frame context through local self-attention. In contrast, MoE blocks incorporate both local and global context. We first apply frame-wise multi-head self-attention to obtain locally updated tokens:
\begin{equation}
    (\,f_n', \:t_n', \:P_n')
    = \mathrm{MultiHeadAttn}\left(\mathbf{T}_n^{(\ell-1)}\right),
    \label{eq:local}
\end{equation}
where $f_n'$ is the updated [CLS] token, $t_n'$ are the FT tokens, and $P_n'$ are the updated patch tokens for frame $n$.

\noindent\textbf{{Global attention over FT tokens (implementation detail).}}
After the local self-attention in Eq.~(\ref{eq:local}), we obtain
$f_n' \in \mathbb{R}^{C}$, $t_n' \in \mathbb{R}^{H \times C}$, and $P_n' \in \mathbb{R}^{M \times C}$ for each frame $n$,
where $f_n'$ is the frame-level $[\mathrm{CLS}]$ token, $t_n'$ are the $H$ FT tokens, and $P_n'$ are the patch tokens.
To inject cross-frame context while keeping the [CLS] token unchanged (so that the core frame-level representative is not altered by global mixing), we exclude $f_n'$ and build a global token sequence by concatenating FT tokens and patch tokens across $F$ frames:
\begin{equation}
    U=\operatorname{concat}\!\big(\,\{[t_n';\,P_n']\}_{n=1}^{F}\big)
    \in \mathbb{R}^{F(H+M)\times C}.
\end{equation}
We then apply global multi-head self-attention over $U$.
Specifically, we compute the projected queries, keys, and values as:
\begin{equation}
    Q = U W_Q,\quad K = U W_K,\quad V = U W_V,
\end{equation}
where $W_Q,W_K,W_V \in \mathbb{R}^{C\times C}$ (implemented in a multi-head manner with head dimension $d$).
The global attention output is given by:
\begin{equation}
    \hat{U}=\mathrm{softmax}\!\left(\frac{QK^\top}{\sqrt{d}}\right)V
    \in \mathbb{R}^{F(H+M)\times C}.
\end{equation}
In a minibatch, this global attention is computed \emph{independently for each video sample}, i.e., tokens from different videos do not attend to each other.

\smallskip
\noindent\textbf{Updating FT tokens only.}
We reshape $\hat{U}$ back to per-frame outputs $\hat{u}_n \in \mathbb{R}^{(H+M)\times C}$ and update only the FT tokens via a residual:
\begin{equation}
    \bar{t}_n = \mathrm{LN}\!\left(t_n' + \hat{u}_n[1{:}H]\right)\in\mathbb{R}^{H\times C},
\end{equation}
while keeping the patch tokens unchanged in this global step, i.e., $\bar{P}_n=P_n'$.
Moreover, updating only FT tokens restricts the pathway through which global information enters the frame representation, preventing direct injection into patch tokens and thereby preserving CLIP's pretrained patch-level semantics.
Within a video, interactions between unrelated frames are further moderated by the \emph{content-based} nature of attention, which learns to assign low weights to semantically irrelevant pairs.

\smallskip
\noindent\textbf{Complexity (layer-wise and branch-wise).}
The global attention over $U$ costs $O\!\left((F(H{+}M))^2 C\right)$ \emph{per video and per layer}.
Importantly, we do \emph{not} apply this global attention to all Transformer layers.
Instead, we deliberately apply global attention over frames only to a small subset of MoE-enabled layers in the
\emph{visual} encoder (e.g., the top $L_G^V{=}2$ backbone layers), and we do not apply global attention over frames
in the text branch ($L_G^T{=}0$).
This design injects cross-frame context where we expand capacity with MoE, while keeping the overall compute overhead small.

Let $S = F(H{+}M)$ denote the length of the global sequence $U$.
Then, the total additional cost of global attention over a minibatch of size $B$ is:
\begin{equation}
\begin{aligned}
C_{\mathrm{global}}
&= O(B \cdot L_G^v \cdot S^2 \cdot C) \\
&= O\!\left(B \cdot L_G^v \cdot (F(H+M))^2 \cdot C\right).
\end{aligned}
\end{equation}
and $\mathcal{C}_{\text{global}}^T = 0$ for the text branch.
Therefore, compared to applying global-over-frames at every layer of both encoders, our design reduces the added cost
proportionally to the small ratio $L_G^V/(L_V{+}L_T)$.

\begin{figure}[!t]
  \raggedright % 그림+캡션을 왼쪽 정렬
  \safeincludegraphics[width=1\linewidth]{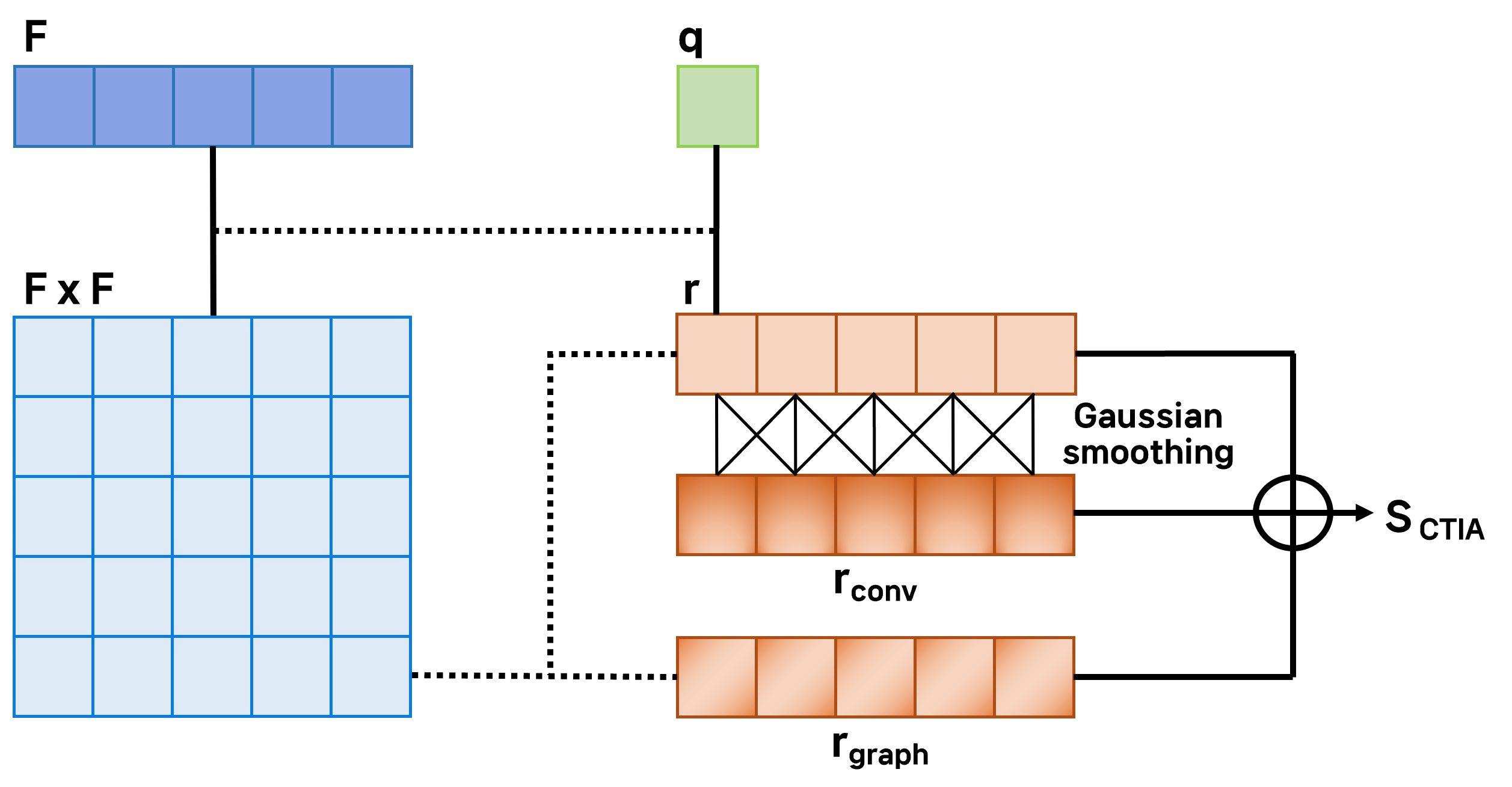}
\caption{CTIA (Cross-Temporal Interaction \& Aggregation) computes the base similarity \(r\) from the sentence vector \(\mathbf{q}\) and frame vectors \(\mathbf{F}\), then applies Gaussian smoothing to \(r\) to obtain \(r_{\text{conv}}\) and generates \(r_{\text{graph}}\) through propagation on the frame--frame similarity graph. It finally fuses these three signals to produce the frame--sentence score \(S_{f,t}\).}
  \label{fig:fig4}
\end{figure}

\subsection{Cross-Temporal Interaction \& Aggregation (CTIA)} \label{Sec:CTIA}

As designed in the previous section, the MoE layers enhance representational capacity by enabling expert specialization at the frame and token levels. However, while the vision-side MoE effectively refines each frame independently, it still lacks an explicit mechanism for inter-frame alignment and temporal consolidation, which are essential for building a coherent temporal stream. To address this limitation, we introduce Cross-Temporal Interaction and Aggregation (CTIA), which injects sentence-conditioned signals into frames and aggregates them into a unified decision score. 
To explicitly connect framewise similarities to the final retrieval decision, we formulate CTIA as a sentence-conditioned aggregation function $S_{\text{CTIA}}=\mathrm{Agg}(\mathbf{s},\mathbf{F})$ that transforms the framewise evidence $\mathbf{s}$ into weights over frames and produces a decision-level score used in $S_{\text{final}}$.

CTIA is built upon three complementary relevance signals: (i) the raw query-conditioned evidence $\mathbf{r}$, (ii) its locally smoothed variant $\mathbf{r}_{\mathrm{conv}}$ that enforces short-range temporal coherence, and (iii) the graph-propagated signal $\mathbf{r}_{\mathrm{graph}}$ that transfers relevance across semantically similar but temporally distant frames. We fuse these signals linearly for interpretability and stable optimization; ablations that remove each term (via $\beta$-term configurations) are provided in Table~\ref{tab:ctia_beta_ablation_r5}.

We first compute a framewise cosine similarity vector $\mathbf{s} = (s_1, s_2, \dots, s_F) \in \mathbb{R}^{F}$ between each frame representation $f_n$ and the sentence embedding $q$ (i.e., $q=e_{[\text{EOS}]}$). 
Based on this framewise evidence, CTIA (Fig.~\ref{fig:fig4}) derives the relevance signals $\mathbf{r}$, $\mathbf{r}_{\mathrm{conv}}$, and $\mathbf{r}_{\mathrm{graph}}$, and produces the final frame weights $\mathbf{w} = (w_1, \dots, w_F) \in \mathbb{R}^{F}$ through four stages: sharpening, local Gaussian smoothing, non-local frame-graph propagation, and learnable fusion.

\smallskip
\noindent\textbf{(1) Sharpening.}
We apply a softmax along the frame axis to obtain per-frame relevance $\mathbf{r} \in \mathbb{R}^{F}$:
\begin{equation}
    \mathbf{r} = \mathrm{softmax}\!\left(\frac{\mathbf{s}}{\tau_s}\right),
\end{equation}
where $\tau_s$ is the temperature parameter.

\smallskip
\noindent \textbf{(2) Local Gaussian smoothing.} We smooth $\mathbf{r}$ with a normalized 1D Gaussian kernel $k \in \mathbb{R}^{K}$ (odd $K$, width $\sigma$) to encode short-range temporal coherence. The kernel is defined as:
\begin{equation}
    k[u] \propto \exp\!\left(-\frac{1}{2}\left(\frac{u}{\sigma}\right)^{2}\right).
\end{equation}
The locally smoothed relevance vector $\mathbf{r}_{\mathrm{conv}}$ is computed by:
\begin{equation}
    \mathbf{r}_{\mathrm{conv}}=\mathbf{r} * k\;\in\mathbb{R}^{F}.
\end{equation}
This operation smooths out fluctuations in frame relevance by averaging neighboring frames. However, local smoothing alone cannot connect repeated or visually similar scenes that are temporally distant, motivating a complementary non-local step.

\smallskip
\noindent \textbf{(3) Non-local frame-graph propagation.} To complement the limitations of local smoothing, we propagate relevance over a content-aware frame graph. Using the frame feature $\mathbf{F} \in \mathbb{R}^{F \times C}$, we compute a frame-to-frame affinity matrix $\mathbf{A} \in \mathbb{R}^{F \times F}$ as:
\begin{equation}
    \mathbf{A} = \mathrm{softmax}\!\left(\frac{\mathbf{F}\mathbf{F}^{\top}}{\tau_g}\right),
\end{equation}
where $\tau_g$ is a temperature parameter. To prevent over-smoothing, we add a self-loop with a balancing coefficient $\gamma \in [0,1]$:
\begin{equation}
    \mathbf{A} \leftarrow \gamma I + (1-\gamma)\mathbf{A},
\end{equation}
where $I$ denotes the identity matrix. The non-local relevance $\mathbf{r}_{\mathrm{graph}}$ is then obtained by propagating $\mathbf{r}$ over the graph:
\begin{equation}
    \mathbf{r}_{\mathrm{graph}} = \mathbf{A}\,\mathbf{r}.
\end{equation}
Unlike local convolution, graph propagation connects frames that are \emph{semantically similar} regardless of temporal distance, reinforcing repeated or thematically related content while the self-loop term $\gamma I$ prevents over-diffusion.

% (Requires xcolor) \usepackage{xcolor}

% (Requires xcolor) \usepackage{xcolor}

\smallskip
\noindent \textbf{(4) Learnable fusion and final weighting.}
We combine the three relevance signals using learnable weights
$\boldsymbol{\beta} = \mathrm{ReLU}([\beta_r,\beta_c,\beta_g])$ and a fusion scale $\lambda_g \ge 0$.
The fused relevance vector $\mathbf{u}$ is:
\begin{equation}
    \mathbf{u}
    = \beta_r\,\mathbf{r} \;+\; \beta_c\,\mathbf{r}_{\mathrm{conv}}
    \;+\; \beta_g\,{\lambda_g}\,\mathbf{r}_{\mathrm{graph}}.
\end{equation}
We then compute the final frame weights $\mathbf{w} \in \mathbb{R}^{F}$ using a temperature-controlled softmax:
\begin{equation}
    \mathbf{w}
    = \mathrm{softmax}\!\left(\frac{\mathbf{u}}{\tau_w}\right).
\end{equation}
Given the framewise similarity vector $\mathbf{s} = (s_1, s_2, \dots, s_F)$, the CTIA score $S_{\text{CTIA}}$ is computed as a weighted sum:
\begin{equation}
    S_{\text{CTIA}}
    = \mathbf{w}^{\top}\mathbf{s}
    = \sum_{i=1}^{F} w_i\, s_i.
\end{equation}

Following CLIP4Clip~\cite{CLIP4Clip}, we use a temporal encoder to aggregate the frame features into a video feature $\mathbf{v}$. We then compute the cosine similarity between the video representation $\mathbf{v}$ and the sentence representation $e_{[\text{EOS}]}$ to obtain the baseline text–video score, which we denote as $S_{\text{base}}$. The final retrieval score $S_{\text{final}}$ is computed as the average of the video-level similarity and our CTIA similarity:
\begin{equation}
    S_{\text{final}}
    = \frac{1}{2}\left(S_{\text{base}} + S_{\text{CTIA}}\right).
\end{equation}

In our default setting, we optimize the entire framework end-to-end using the objective in Sec.~\ref{Sec:3E}.

% ====== MSR-VTT-9K (clean layout) ======
\begin{table*}[!t]
\centering
\small
\caption{\textbf{Retrieval performance on the MSR-VTT-9K dataset.} 
The symbol * indicates the use of the DSL~\cite{DSL} post-processing operation in the corresponding methods, 
while the symbol ** indicates that QB-Norm~\cite{QBNorm} is additionally applied, 
and the symbol † denotes the results obtained from the source code provided by the authors. 
Gray-shaded entries indicate results taken from large-scale foundation models with substantially different pretraining scale/objectives; for InternVideo2, only R@1 is reported in the source table, and other metrics are not available (shown as ``--'').}\label{tab:msrvtt}
\ra{1.1}\setlength{\tabcolsep}{5.5pt}
\begin{threeparttable}
\resizebox{\textwidth}{!}{%
\begin{tabular}{lccccc|ccccc}
\toprule
\multirow{2}{*}{\textbf{Method}} &
\multicolumn{5}{c|}{\textbf{Text $\Rightarrow$ Video}} &
\multicolumn{5}{c}{\textbf{Video $\Rightarrow$ Text}} \\
\cmidrule(lr){2-6}\cmidrule(lr){7-11}
& R@1$\uparrow$ & R@5$\uparrow$ & R@10$\uparrow$ & MdR$\downarrow$ & MnR$\downarrow$
& R@1$\uparrow$ & R@5$\uparrow$ & R@10$\uparrow$ & MdR$\downarrow$ & MnR$\downarrow$ \\
\midrule
\multicolumn{11}{l}{\emph{CLIP-ViT-B/32}} \\
CLIP4Clip\textsuperscript{\dag}~\cite{CLIP4Clip} 
& 44.0 & 72.0 & 81.6 & 2.0 & 16.0
& 43.9 & 70.6 & 79.8 & 2.0 & 12.0 \\
CAMoE~\cite{DSL} 
& 44.6 & 72.6 & 81.8 & 2.0 & 13.3
& 45.1 & 72.4 & 83.1 & 2.0 & 10.0 \\
CAMoE*~\cite{DSL} 
& 47.3 & 74.2 & 84.5 & 2.0 & 11.9
& 49.1 & 74.3 & 84.3 & 2.0 & 9.9 \\
CLIP2Video~\cite{CLIP2Video} 
& 45.6 & 72.6 & 81.7 & 2.0 & 14.6
& 43.5 & 72.3 & 82.1 & 2.0 & 10.2 \\
CenterCLIP~\cite{CenterCLIP}           
& 44.2 & 71.6 & 82.1 & 2.0 & 15.1
& 42.8 & 71.7 & 82.2 & 2.0 & 10.9 \\
X-Pool~\cite{XPool}               
& 46.9 & 72.8 & 82.2 & 2.0 & 14.3
& 44.4 & 73.3 & 84.0 & 2.0 &  9.0 \\
TS2-Net~\cite{TS2Net}               
& 47.0 & 74.5 & 83.8 & 2.0 & 13.0
& 45.3 & 74.1 & 83.7 & 2.0 &  9.2 \\
X-CLIP~\cite{XCLIP}                 
& 46.1 & 73.0 & 83.1 & 2.0 & 13.2
& 46.8 & 73.3 & 84.0 & 2.0 &  9.1 \\
QB-Norm**~\cite{QBNorm}   
& 47.2 & 73.0 & 83.0 & 2.0 & --
& --  & --  & --  & --  & --  \\
DRL~\cite{DRL}                 
& 47.4 & 74.6 & 83.8 & 2.0 & --
& --  & --  & --  & --  & --  \\
CLIP-VIP~\cite{CLIPViP}     
& 46.5 & 72.1 & 82.5 & -- & --
& --  & --  & --  & -- & -- \\
STAN~\cite{STAN}               
& 46.9 & 72.8 & 82.8 & 2.0 & --
& --  & --  & --  & -- & -- \\
UATVR~\cite{UATVR}             
& 47.5 & 73.9 & 83.5 & 2.0 & 12.3
& 46.9 & 73.8 & 83.8 & 2.0 & 8.6 \\
UATVR*~\cite{UATVR}              
& 49.8 & 76.1 & 85.5 & 2.0 & 12.9
& 51.1 & 74.8 & 85.1 & 1.0 & 8.3 \\
ProST~\cite{ProST}             
& 48.2 & 74.6 & 83.4 & 2.0 & 12.4
& 46.3 & 74.2 & 83.2 & 2.0 & 8.7 \\
Prompt Switch~\cite{PromptSwitch}    
& 47.8 & 73.9 & 82.2 & -- & 14.1
& 46.0 & 74.3 & \textbf{84.8} & -- & \textbf{8.5} \\
TeachCLIP~\cite{TeachCLIP}    
& 46.8 & 74.3 & -- & -- & --
& --  & --  & --  & -- & -- \\
EERCF~\cite{EERCF}             
& 47.8 & 74.1 & 84.1 & --  & --
& 44.7 & 74.2 & 83.9 & --  & -- \\
LSDO~\cite{LSDO}      
& 49.1 & 75.2 & 84.2 & 2.0 & 12.0
& --  & --  & --  & -- & -- \\
GLSCL~\cite{GLSCL}      
& 48.1 & 73.9 & 82.3 & 2.0 & 13.8
& 47.1 & 73.0 & 83.2 & 2.0 & 10.3 \\
GLSCL**~\cite{GLSCL}      
& 50.0 & 73.7 & 82.8 & 1.5 & 14.2
& 49.4 & 74.5 & 83.9 & 2.0 & 9.6 \\
\textcolor{gray}{InternVideo2$_{\text{s2-6B}}$~\cite{InternVideo2}}
& \textcolor{gray}{62.8} & \textcolor{gray}{--} & \textcolor{gray}{--} & \textcolor{gray}{--} & \textcolor{gray}{--}
& \textcolor{gray}{60.2} & \textcolor{gray}{--} & \textcolor{gray}{--} & \textcolor{gray}{--} & \textcolor{gray}{--} \\
\rowcolor{gray!20}
TAME      
& 48.0 & 74.1 & 81.6 & 2.0 & 14.2
& 46.9 & 72.7 & 82.1 & 2.0 & 10.6 \\
\rowcolor{gray!20}
TAME*      
& \textbf{51.1} & \textbf{75.8} & \textbf{84.5} & \textbf{1.0} & \textbf{10.9}
& \textbf{52.1} & \textbf{76.5} & 83.3 & \textbf{1.0} & \textbf{8.5} \\
\bottomrule
\end{tabular}}
\end{threeparttable}
\end{table*}

\subsection{Objective} \label{Sec:3E}
Given a minibatch $\{(v_i, q_i)\}_{i=1}^{B}$, where $v_i$ denotes the $i$-th video and $q_i$ the corresponding text query, we construct a similarity matrix $\mathbf{S} \in \mathbb{R}^{B \times B}$, where each entry represents the final video–text similarity:
\begin{equation}
    S_{ij} = S_{\text{final}}(v_i, q_j).
\end{equation}
We then optimize the bidirectional InfoNCE losses~\cite{InfoNCE}:
\begin{equation}
\begin{aligned}
    \mathcal{L}_{q \rightarrow v}
    &=
    -\frac{1}{B}\sum_{i=1}^{B}
        \log\frac{\exp(S_{ii}/\tau)}
                 {\sum_{j=1}^{B}\exp(S_{ij}/\tau)}, \\
    \mathcal{L}_{v \rightarrow q}
    &=
    -\frac{1}{B}\sum_{i=1}^{B}
        \log\frac{\exp(S_{ii}/\tau)}
                 {\sum_{j=1}^{B}\exp(S_{ji}/\tau)}, \\
    \mathcal{L}_{vq}
    &=
    \frac{1}{2}\big(\mathcal{L}_{q\rightarrow v}
        +\mathcal{L}_{v\rightarrow q}\big).
\end{aligned}
\end{equation}
Here, $\tau$ is the temperature hyperparameter for scaling.

Additionally, we apply a batch-hard margin loss~\cite{hermans2017defense, liu2019trajectory} to further separate each positive pair from its most confusing negative. For each anchor $i$, we denote the positive similarity as $p_i = S_{ii}$, and identify the hardest negative as $n_i = \max_{j \neq i} S_{ij}$. The loss penalizes cases where the hardest negative becomes closer than the positive by a margin $m$:
\begin{equation}
    \mathcal{L}_{\text{hard}}
    =
    \frac{1}{B}\sum_{i=1}^{B}
    \big[m + n_i - p_i\big]_+,
    \qquad [x]_+ = \max(x, 0).
\end{equation}
Here, $m>0$ is the margin hyperparameter that enforces a minimum separation between the positive and its hardest negative.

Finally, the overall training objective combines the bidirectional InfoNCE loss, the MoE load-balancing regularizers, and the hard negative margin loss: \begin{equation} 
    \mathcal{L}_{\text{total}} 
    = \mathcal{L}_{vq} \;+\; \lambda \big( \mathcal{L}^{T}_{\text{aux}} + \mathcal{L}^{V}_{\text{aux}} \big) \;+\; \mathcal{L}_{\text{hard}}. 
\end{equation} 
Here, $\lambda$ is a hyperparameter.

% ====== Unified Table (DiDeMo / MSVD / LSMDC / ActivityNet) ======
\begin{table*}[!t]
\centering
\begingroup
\caption{Text-to-video retrieval results on \textbf{DiDeMo}, \textbf{MSVD}, \textbf{LSMDC}, and \textbf{ActivityNet}. We report standard recall metrics (R@1/R@5/R@10; higher is better) and mean rank (MnR; lower is better) under the official evaluation protocols of each benchmark. Missing metrics that are not reported in the original papers are marked as ``-''. The symbol * indicates the use of the DSL~\cite{DSL}.} \label{tab:unified_retrieval}
\setlength{\tabcolsep}{4pt}
\scriptsize
\resizebox{\textwidth}{!}{%
\begin{tabular}{l | cccc | cccc | cccc | cccc}
\toprule
\multirow{2}{*}{\textbf{Method}} 
& \multicolumn{4}{c|}{\textbf{DiDeMo}} 
& \multicolumn{4}{c|}{\textbf{MSVD}} 
& \multicolumn{4}{c|}{\textbf{LSMDC}} 
& \multicolumn{4}{c}{\textbf{ActivityNet}} \\
\cmidrule(lr){2-5}\cmidrule(lr){6-9}\cmidrule(lr){10-13}\cmidrule(lr){14-17}
& R@1$\uparrow$ & R@5$\uparrow$ & R@10$\uparrow$ & MnR$\downarrow$
& R@1$\uparrow$ & R@5$\uparrow$ & R@10$\uparrow$ & MnR$\downarrow$
& R@1$\uparrow$ & R@5$\uparrow$ & R@10$\uparrow$ & MnR$\downarrow$
& R@1$\uparrow$ & R@5$\uparrow$ & R@10$\uparrow$ & MnR$\downarrow$ \\
\midrule
CLIP4Clip~\cite{CLIP4Clip}         & 42.8 & 68.5 & 79.2 & 18.9 & 45.2 & 75.5 & 84.3 & 10.3 & 22.6 & 41.0 & 49.1 & 61.0 & 40.5 & 72.4 & 83.6 & 7.5 \\
CenterCLIP~\cite{CenterCLIP}       &  -   &  -   &  -   &  -   &  -   &  -   &  -   &  -   & 21.4 & 39.7 & 49.4 & 55.9 & 43.5 & 75.0 & 85.9 & 6.9 \\
CAMoE~\cite{DSL}                 & 43.8 & 71.4 &  -   &  -   & 46.9 & 76.1 &  -   &  9.8 & 22.5 & 42.6 & 50.9 & 56.5 &  -   &  -   &  -   &  -   \\
CLIP2Video~\cite{CLIP2Video}       &  -   &  -   &  -   &  -   & 47.0 & 76.8 & 85.9 &  9.6 &  -   &  -   &  -   &  -   &  -   &  -   &  -   &  -   \\
X-Pool~\cite{XPool}                &  -   &  -   &  -   &  -   & 47.2 & 77.4 & 86.0 &  \textbf{9.3} & \textbf{25.2} & \textbf{43.7} & \textbf{53.5} & \textbf{53.2} &  -   &  -   &  -   &  -   \\
TS2-Net~\cite{TS2Net}              & 41.8 & 71.6 & 82.0 & 14.8 &  -   &  -   &  -   &  -   & 23.4 & 42.3 & 50.9 & 56.9 & 41.0 & 73.6 & 84.5 & 8.4 \\
X-CLIP~\cite{XCLIP}                & 45.2 & 74.0 &  -   & 14.6 & 47.1 & 77.8 &  -   &  9.5 & 23.3 & 43.0 &  -   & 56.0 & 44.3 & 74.1 &  -   & 7.9 \\
CLIP-VIP~\cite{CLIPViP}            & 40.6 & 70.4 & 79.3 & 15.1 &  -   &  -   &  -   &  -   &  -   &  -   &  -   &  -   &  -   &  -   &  -   &  -   \\
UATVR~\cite{UATVR}                 & 43.1 & 71.8 & 82.3 & 15.1 & 46.0 & 76.3 & 85.1 & 10.4 &  -   &  -   &  -   &  -   &  -   &  -   &  -   &  -   \\
ProST~\cite{ProST}                 & 44.9 & 72.7 & 82.7 & 13.7 &  -   &  -   &  -   &  -   & 24.1 & 42.5 & 51.6 & 54.6 &  -   &  -   &  -   &  -   \\
Prompt Switch~\cite{PromptSwitch}  &  -   &  -   &  -   &  -   & 47.1 & 76.9 & \textbf{86.1} &  9.5 & 23.1 & 41.7 & 50.5 & 56.8 &  -   &  -   &  -   &  -   \\
TeachCLIP~\cite{TeachCLIP}         & 43.7 & 71.2 &  -   &  -   & 47.4 & 77.3 &  -   &  -   &  -   &  -   &  -   &  -   & 42.2 & 72.7 &  -   &  -   \\
EERCF~\cite{EERCF}                 &  -   &  -   &  -   &  -   &  -   &  -   &  -   &  -   &  -   &  -   &  -   &  -   &  43.1 & 74.5 & - & -  \\
LSDO~\cite{LSDO}                   & 44.7 & 71.1 & 81.8 & 14.5 & 44.6 & 75.4 & 84.8 & 10.3 &  -   &  -   &  -   &  -   &  -   &  -   &  -   &  -   \\
GLSCL~\cite{GLSCL}                 & 46.7 & 75.0 & 82.9 & 13.0 & 47.5 & 76.3 & \textbf{86.1} &  9.8 & 23.4 & 42.4 & 49.8 & 56.2 & 40.6 & 71.8 & 84.1 & 7.1 \\
\midrule
\rowcolor{gray!20}
TAME                                & 45.5 & 72.9 & 80.8 & 14.8 & 46.6 & 76.8 & 85.4 &  9.9 & 24.1 & 41.4 & 51.3 & 57.6 & 42.7 & 73.0 & 84.6 & 7.6 \\
\rowcolor{gray!20}
TAME~*                              & \textbf{49.4} & \textbf{75.1} & \textbf{83.0} & \textbf{12.9}
                                   & \textbf{48.7} & \textbf{77.9} & 85.9 & 10.4
                                   & 23.4 & 43.3 & 51.9 & 56.0
                                   & \textbf{48.9} & \textbf{77.0} & \textbf{87.0} & \textbf{6.4} \\
\bottomrule
\end{tabular}%
}
\endgroup
\end{table*}

\section{Experiments}
\subsection{Datasets}
We evaluate on five widely used text--video retrieval benchmarks and, unless otherwise noted, follow standard splits and protocols. 
\textbf{MSR-VTT}~\cite{msrvtt} contains 10{,}000 web video clips, each annotated with 20 sentences; following~\cite{mtv}, we use 9{,}000 clips for training and report results on the standard 1k-A test split. 
\textbf{DiDeMo}~\cite{didemo} includes 10{,}000 Flickr videos with 40{,}000 sentence descriptions, and we concatenate all descriptions of a video into a single paragraph and perform paragraph-to-video retrieval. 
\textbf{MSVD}~\cite{msvd} comprises 1{,}970 YouTube videos with roughly 40 crowd-sourced captions per video, and we adopt the 1{,}200/100/670 train/val/test split.
\textbf{LSMDC}~\cite{lsmdc} is a movie-sourced benchmark with 118{,}081 short clips (2--30 seconds) collected from 202 movies; following~\cite{CLIP4Clip},~\cite{XPool}, we use 7{,}408 videos for validation and 1{,}000 videos for testing.
\textbf{ActivityNet}~\cite{activitynet} contains around 20{,}000 YouTube videos; following prior videoparagraph retrieval protocols~\cite{hse},~\cite{mtv}, we concatenate all descriptions of a video into a single paragraph and evaluate on the 'val1' split.

\subsection{Evaluation}
We report results for both text-to-video (T$\to$V) and video-to-text (V$\to$T) retrieval. 
We evaluate using Recall@K (R@1, R@5, R@10), Median Rank (MdR), and Mean Rank (MnR). 
R@K is the fraction of queries whose first correct match appears within the top-$K$, and MdR/MnR are the median and mean rank, respectively, of that first correct match.
\subsection{Implementation details.}
Following CLIP4Clip~\cite{CLIP4Clip}, we initialize the text and video encoders with the pretrained CLIP (ViT-B/32) weights~\cite{CLIP}. 
We optimize with Adam~\cite{Adam} and decay the learning rate using a cosine schedule~\cite{sgdr}. The initial learning rate is set to \(1\times10^{-7}\) for the text and video encoders (including linear projections), and to \(1\times10^{-4}\) for the other modules.
For MSR-VTT, MSVD and LSMDC we set the maximum token length to 32 and the maximum number of frames per video to 12. For DiDeMo and ActivityNet, we use 64 tokens and 64 frames. The batch size and number of training epochs are 128 and 5 for MSR-VTT, 300 and 5 for MSVD, LSMDC 64 and 20 for DiDeMo and 64 and 25 for ActivityNet.  For MoE-Enhanced CLIP, we replace the FFNs of the last two Transformer layers (\(\{11,12\}\)) in both encoders with sparse MoE blocks using \(N_E{=}4\) experts and Top-\(K\) routing (\(K{=}2\) for text and \(K{=}1\) for vision). We set the number of Frame–Temporal (FT) tokens to \(H{=}2\). Experiments are conducted on four NVIDIA RTX A6000 GPUs. On the MSR-VTT 1k-A split, training requires approximately 4 hours for 5 epochs. During inference, computing the query--video similarity matrix takes 20.84 s in total (20.8\,ms per query).

\noindent\textbf{Reporting protocol.}
Unless specified, we report raw similarities. When indicated (e.g., “*”), we apply evaluation-time DSL~\cite{DSL} with temperature \(\tau_{\mathrm{dsl}}\), which reweights similarities via softmax along both the query and video dimensions and combines the two weights into the final score.
Notably, DSL is applied only at inference time and does not affect training.

% ---- Fig5: Routing metrics vs training steps (MoE placements {11,12}) ----
\begin{figure*}[t]
  \centering
  \safeincludegraphics[width=\textwidth]{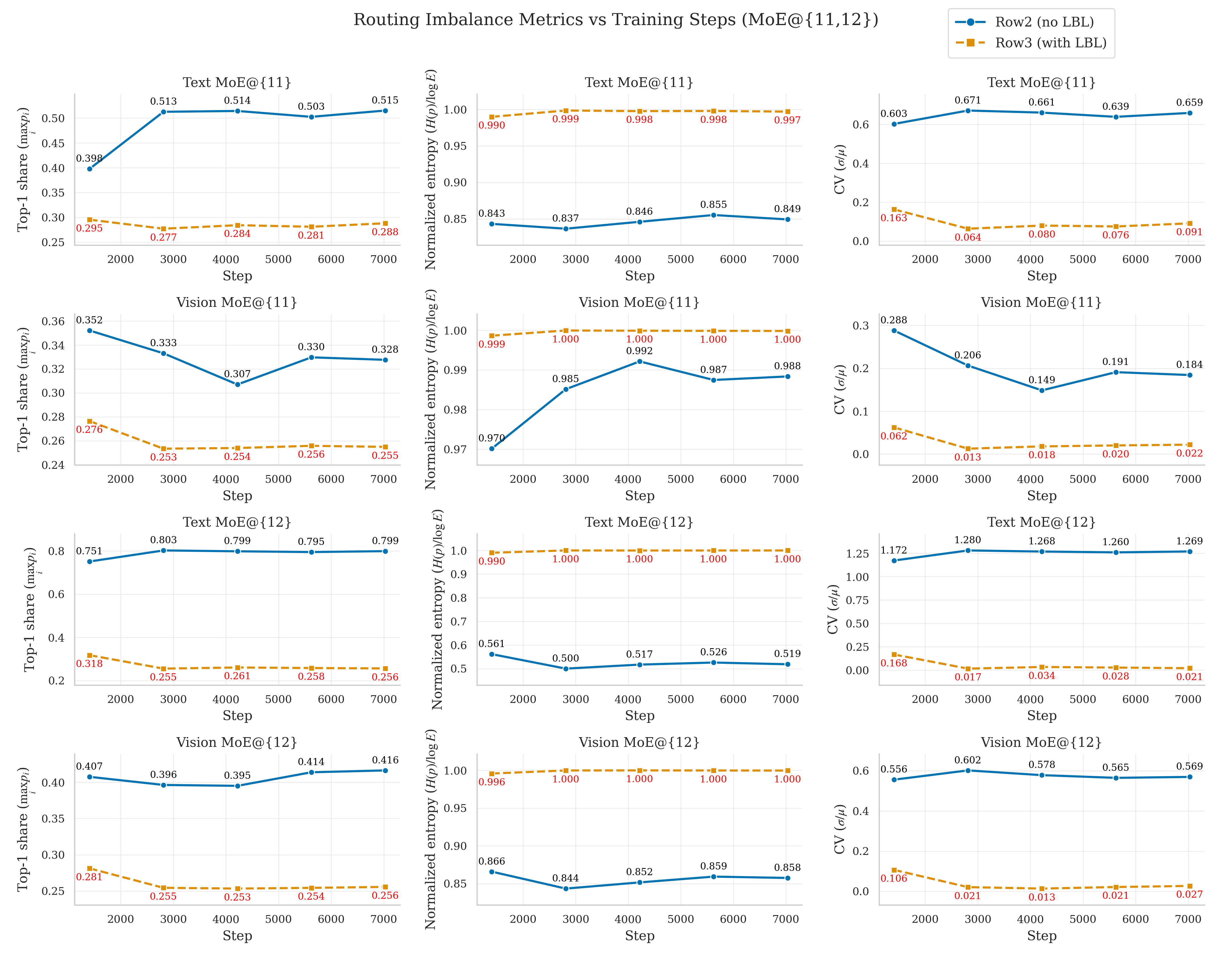}
  \caption{\textbf{Routing distribution metrics over training for MoE placements \(\{11,12\}\).}
    We plot the top-1 share \(\max_i p_i\), normalized entropy \(H(p)/\log E\), and coefficient of variation \(\sigma(p)/\mu(p)\),
    where \(p_i\) denotes the average routing probability assigned to expert \(i\) and \(E\) is the number of experts.
    Without the auxiliary load-balancing loss (Row2), the routing distribution becomes persistently skewed, indicating layer-selective collapse.
    With load-balancing (Row3), the routing distribution remains stable and close to uniform throughout training.}
  \label{fig:moe_metrics_steps_1112}
  \vspace{-2mm}
\end{figure*}

\begin{table}[!t]
\centering
\small
\textcolor{blue}{}
\caption{{Ablation study of the proposed components. We progressively incorporate each module into the same CLIP backbone under identical settings (MoE: sparse upcycling; LBL: load balancing; FT: Frame–Temporal tokens; CFA: cross-frame attention (global attention over FT tokens); CTIA: sentence-conditioned temporal aggregation).}} \label{tab:ablation_incremental}
\ra{1.15}\setlength{\tabcolsep}{5pt}
\resizebox{\linewidth}{!}{%
\begin{threeparttable}
\begin{tabular}{lccccc} % 6 columns: 1 + 3 (R@K) + 2 (MdR,MnR)
\toprule
\multirow{2}{*}{\raisebox{-0.25\totalheight}{\textbf{Configuration}}} & \multicolumn{3}{c}{R@K $\uparrow$} & \multirow{2}{*}{MnR $\downarrow$} \\
\cmidrule(lr){2-4}
& R@1 & R@5 & R@10 & & \\
\midrule
(1) CLIP4Clip baseline                         & 44.0 & 72.0 & 81.6  & 16.0 \\
(2) (1) + MoE                                  & 43.0 & 70.6 & 80.4  & 16.1 \\
(3) (2) + LBL            & 43.8 & 71.1 & 80.0 & 16.0 \\
(4) (3) + FT         & 42.8 & 71.4 & 80.8 & 16.4 \\
(5) (4) + CFA          & 42.9 & 71.2 & 81.2 & 16.1 \\
(6) (5) + CTIA                                 & 47.5 & 72.5 & \textbf{82.0} & \textbf{13.2} \\
(7) (6) + Hard-negative loss               & \textbf{48.0} & \textbf{74.1} & 81.6 & 14.2 \\
\bottomrule
\end{tabular}
\end{threeparttable}
}% resizebox
\end{table}

\subsection{Performance Comparisons}
We compare our model against methods that use CLIP-ViT as the backbone. 
On \textbf{MSR-VTT} (Table~\ref{tab:msrvtt}), our model achieves an improvement of \(\mathbf{+4.0}\) R@1 over the CLIP4Clip~\cite{CLIP4Clip} baseline. 
Compared to \textbf{X-CLIP}, which employs multi-level alignment (video--sentence, video--word, frame--sentence, frame--word), our approach relies only on video–sentence and frame–sentence alignment yet achieves a \(\mathbf{+1.9}\) R@1, underscoring the effectiveness of CTIA-based temporal alignment. 
Applying the DSL post-processing further boosts performance across all metrics.

As reported in Table~\ref{tab:unified_retrieval}, we present results on the \textbf{DiDeMo}, \textbf{MSVD}, \textbf{LSMDC}, and \textbf{ActivityNet} datasets. Our model demonstrates stable generalization across these datasets, yielding consistent improvements over the baseline.
In particular, on \textbf{DiDeMo}, R@1 improves by \(\mathbf{+2.7}\) over the baseline. On \textbf{MSVD}, R@1 improves by \(\mathbf{+1.4}\) over the baseline.
For \textbf{LSMDC}, our model achieves a \(\mathbf{+1.5}\) improvement in R@1 over CLIP4Clip, increasing from 22.6 to 24.1, while also reducing MnR from 61.0 to 57.6.
Furthermore, \textbf{ActivityNet}, a long-form benchmark, shows a +2.2 improvement in R@1 over CLIP4Clip, improving from 40.5 to 42.7. It also yields consistent improvements at higher cutoffs, with R@5 reaching 73.0 and R@10 reaching 84.6.
Finally, we confirm that applying DSL post-processing yields additional performance gains across all four datasets: R@1 increases by \(\mathbf{+3.9}\) on DiDeMo (45.5 \(\rightarrow\) 49.4), by \(\mathbf{+2.1}\) on MSVD (46.6 \(\rightarrow\) 48.7), and by \(\mathbf{+6.2}\) on ActivityNet (42.7 \(\rightarrow\) 48.9). On LSMDC, while R@1 slightly decreases (24.1 \(\rightarrow\) 23.4), high-recall metrics improve (R@5: 41.4 \(\rightarrow\) 43.3; R@10: 51.3 \(\rightarrow\) 51.9) together with a lower MnR (57.6 \(\rightarrow\) 56.0), indicating that DSL provides additional ranking benefits in terms of broader recall.

\subsection{Ablation Study}
\label{subsec:ablations}
We conduct ablations on \textbf{MSR-VTT (1k-A split)} to analyze the contribution of each component. The corresponding results are summarized in Table~\ref{tab:ablation_incremental}.
Specifically, the ablation study evaluates the following components: (a) the CLIP4Clip baseline, (b) \textbf{MoE} in both encoders, (c) \textbf{Load-balancing loss (LBL)} for MoE routing, (d) \textbf{Frame–Temporal (FT) tokens} inserted before each visual MoE layer, (e) \textbf{Cross-frame attention (CFA; global attention over FT tokens}), (f) \textbf{CTIA} for cross-temporal interaction/aggregation, and (g) \textbf{Hard-negative} margin loss. Unless otherwise stated, all settings (backbone, training epochs, optimizer, and data pre/post-processing) are fixed across rows.

% ---- Fig6: Expert ratio bar panels (MoE placements {11,12}) ----
\begin{figure*}[t]
  \centering
  \safeincludegraphics[width=\textwidth]{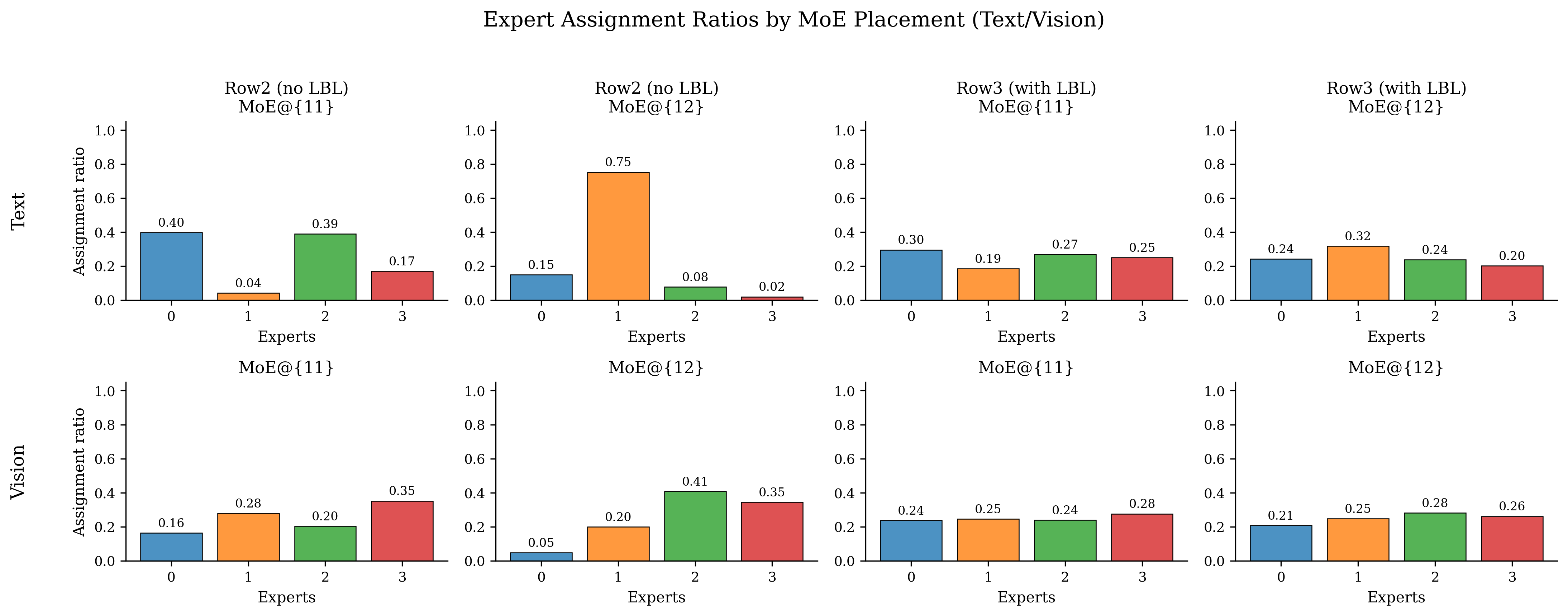}
  \caption{\textbf{Expert assignment ratios for MoE placements \(\{11,12\}\).}
    We visualize the per-expert routing distribution \(p\in\mathbb{R}^{E}\) for both text and vision encoders during the initial fine-tuning stage.
    Row~2 (no LBL) exhibits strong routing concentration, most prominently in the text MoE at \(\{12\}\), whereas Row~3 (with LBL) yields near-uniform utilization across experts.
    This figure reports the corresponding imbalance trajectories over training.}
  \label{fig:expert_ratio}
  \vspace{-2mm}
\end{figure*}

Starting from the CLIP4Clip baseline (Row~1), we first replace a subset of feed-forward layers in both the text and vision encoders with Mixture-of-Experts (MoE) layers (Row~2). Simply inserting MoE without any additional regularization does not yield an immediate improvement in retrieval quality. We interpret this as early evidence that naively adding sparsely activated experts can introduce routing instability and inconsistent specialization across frames and tokens, which hurts retrieval alignment.

To make this instability more explicit, Fig.~\ref{fig:moe_metrics_steps_1112} tracks routing imbalance over training for MoE placements $\{11,12\}$, and Fig.~\ref{fig:expert_ratio} visualizes the expert assignment ratio for each MoE layer as bar charts over experts. We report three imbalance indicators: (i) top-1 share \(\max_i p_i\), which measures the dominance of the most frequently selected expert, (ii) normalized entropy \(H(p)/\log E\) (closer to 1 indicates more uniform routing), and (iii) coefficient of variation (CV) \(\sigma(p)/\mu(p)\). Unless noted otherwise, the MoE layer \(\{12\}\) refers to the top CLIP backbone block and \(\{11\}\) denotes the second-to-top block.
The diagnostics reveal a pronounced layer-selective routing collapse in Row~2, dominated by the text encoder at \(\{12\}\): early in training, a single expert becomes highly dominant (top-1 share \(0.75\sim0.80\)), accompanied by reduced entropy and large dispersion (e.g., \(H(p)/\log E \approx 0.50\sim0.56\), \(\text{CV}\approx 1.17\sim1.28\)). By comparison, the \(\{11\}\) MoE layer is less extreme but still noticeably imbalanced (top-1 \(\approx 0.50\) with \(\text{CV}\approx 0.64\sim0.67\)), indicating that the instability is amplified near the top layer rather than being uniform across MoE placements. On the vision side, the imbalance at \(\{12\}\) is milder (typical top-1 \(\approx 0.40\) with \(H(p)/\log E \approx 0.84\sim0.86\)). Fig.~\ref{fig:expert_ratio} provides a complementary qualitative snapshot: the expert-ratio bar charts reveal that Row~2 (no LBL) assigns a disproportionate fraction of tokens to a single expert in the text MoE at \(\{12\}\), while the remaining experts receive markedly smaller shares. This visual evidence is consistent with the imbalance trends quantified in Fig.~\ref{fig:moe_metrics_steps_1112} (e.g., elevated top-1 share and reduced entropy). Overall, the results indicate that token-wise routing in the text encoder—particularly at the top MoE layer \(\{12\}\)—is a major contributor to the pronounced performance drop observed when MoE is introduced without stabilization/regularization.

To address this, we add a load-balancing loss (LBL) on the expert router (Row~3). LBL encourages more uniform expert utilization and prevents the router from collapsing onto a few experts. With this regularization, performance stabilizes and the distribution of active experts becomes more balanced. In other words, balanced expert assignment is important for maintaining stable text--video alignment.

This stabilization is also confirmed by the routing diagnostics: with LBL (Row~3), routing at the same upper-layer MoE placements \(\{11,12\}\) remains near-uniform throughout training. In particular, both \(\{11\}\) and \(\{12\}\) exhibit \(\text{entropy}_{\text{norm}}\approx0.99\sim1.00\) with low top-1 share and low CV, indicating balanced expert utilization and preventing router collapse. This consistent behavior supports that LBL stabilizes optimization and mitigates the early degradation observed in Row~2.

We note that Row~3 does not yet surpass the dense CLIP4Clip baseline (Row~1), even though routing collapse is effectively mitigated. This is expected: LBL mainly serves as a \emph{stabilizer} that prevents degenerate expert utilization and ensures consistent optimization, rather than directly boosting retrieval alignment on its own. In the early fine-tuning stage on MSR-VTT, MoE also introduces additional degrees of freedom (router and expert specialization), which can delay immediate gains.
Accordingly, the severe early degradation in Row~2 is best explained by unstabilized routing collapse, whereas Row~3 should be viewed as a necessary intermediate step that makes the MoE backbone reliably trainable. In particular, without LBL the top-layer text router quickly saturates to a single expert, leading to highly non-smooth gradients and unstable token-level alignment early in training (Fig.~\ref{fig:moe_metrics_steps_1112}--\ref{fig:expert_ratio}).

Building on this stabilized MoE backbone, we next assess the incremental impact of temporal modeling (Rows~4--5).
We then insert Frame--Temporal (FT) tokens  before each visual MoE layer (Row~4) so that each frame representation is augmented with lightweight temporal tokens. However, using FT tokens alone (Row~4) does not materially improve retrieval quality, since FT tokens enrich each frame in isolation but does not explicitly exchange information across frames. To address this limitation, we add cross-frame attention (CFA) (Row~5), which allows these tokens to exchange information across frames. CFA stabilizes performance and prevents further degradation in rank metrics, indicating that explicit inter-frame exchange is important for temporal modeling.

Row~6 introduces our Cross-Temporal Interaction \& Aggregation (CTIA) module, which aggregates sentence--frame similarity over time and produces a video-level score. At this stage we observe a clear improvement over the baseline: the model ranks relevant videos higher on average. Taken together with Rows~4--5, these results suggest that strengthening frame-level representations (via FT and CFA) is helpful but not sufficient on its own; the retrieval model benefits most when those per-frame signals are subsequently organized across time. CTIA performs this temporal organization by aggregating and propagating evidence across frames to produce a more coherent video-level match. Rather than relying only on a single pooled embedding, it highlights the frames that are most relevant to the text query and links them into a consistent trajectory. Finally, Row~7 adds a hard negative margin loss, which penalizes cases where non-matching videos are scored too close to the ground-truth match. This further sharpens retrieval by explicitly separating hard distractors, especially among visually similar candidates.

\begin{table}[!t]
\caption{{Analysis of the effect of MoE layer placement by varying only the encoder layers in which FFNs are upcycled to MoE, while fixing \(E\), top-$K$ routing, and all other settings.}} \label{tab:ablation_layer_placement_compact}
\centering
\scriptsize
\ra{1.1}\setlength{\tabcolsep}{4pt}
\resizebox{\linewidth}{!}{%
\begin{tabular}{lccccc}
\toprule
\textbf{Layers} & R@1 $\uparrow$ & R@5 $\uparrow$ & R@10 $\uparrow$ & MnR $\downarrow$ \\
\midrule
\{12\}                 & 47.9 & 73.7 & \textbf{82.2} & 14.2 \\
\{9,11\}               & 47.4 & 73.7 & 81.9          & 14.1          \\
\{11,12\} (def.)       & \textbf{48.0} & \textbf{74.1} & 81.6          & 14.2 \\
\{8,10,12\}            & 46.4 & 72.0 & 81.7          & \textbf{13.4}          \\
\{10,11,12\}           & 45.9 & 73.3 & 81.5          & 14.0          \\
\{7--12\}              & 45.8 & 73.6 & 82.1          & 13.9          \\
\bottomrule
\end{tabular}
}% end resizebox
\end{table}

\begin{table}[!t]
\caption{{Effect of the number of MoE experts, measured by varying only \(E\) in the upcycled FFNs, while keeping the MoE placement, top-$K$ routing, and all other settings fixed.}} \label{tab:ablation_joint_equal_experts}
\centering
\scriptsize
\ra{1.1}\setlength{\tabcolsep}{4pt}
\resizebox{\linewidth}{!}{%
\begin{threeparttable}
\begin{tabular}{cccccc}
\toprule
$N_E$ & R@1 $\uparrow$ & R@5 $\uparrow$ & R@10 $\uparrow$ & MdR $\downarrow$ & MnR $\downarrow$ \\
\midrule
2  & 46.7 & 72.2 & \textbf{82.7} & 2.0 & 13.9 \\
4  & \textbf{48.0} & \textbf{74.1} & 81.6 & 2.0 & 14.2 \\
6  & 46.8 & 72.3 & 82.0 & 2.0 & 14.3 \\
8  & 46.4 & 73.1 & 81.7 & 2.0 & \textbf{13.6} \\
\bottomrule
\end{tabular}
\end{threeparttable}
}% end resizebox
\end{table}

\begin{table}[!t]
\caption{{Effect of the number of Frame--Temporal Tokens. The number of tokens inserted per frame is varied to control temporal context modeling, while keeping the CLIP backbone and other modules fixed.}}
\label{tab:ablation_ft_count}
\centering
\scriptsize
\ra{1.1}\setlength{\tabcolsep}{4pt}
\resizebox{\linewidth}{!}{%
\begin{threeparttable}
\begin{tabular}{ccccccc}
\toprule
\textbf{FT} & R@1 $\uparrow$ & R@5 $\uparrow$ & R@10 $\uparrow$ & MdR $\downarrow$ & MnR $\downarrow$ \\
\midrule
0 & 47.4 & 73.7 & 81.4 & 2.0 & \textbf{13.4} \\
1 & 47.1 & 72.6 & 81.5 & 2.0 & 14.4 \\
2 & \textbf{48.0} & \textbf{74.1} & 81.6 & 2.0 & 14.2 \\
3 & 47.1 & 73.0 & \textbf{82.1} & 2.0 & 13.7 \\
\bottomrule
\end{tabular}
\end{threeparttable}
}% end resizebox
\end{table}

\begin{table}[t]
\centering
\normalsize   
\setlength{\tabcolsep}{3.5pt}
\caption{CTIA $\beta$-term ablation on MSR-VTT (1k-A), evaluating the contribution of each relevance component ($\mathbf{r}$, $\mathbf{r}_{\mathrm{conv}}$, $\mathbf{r}_{\mathrm{graph}}$) and their combinations.}
\label{tab:ctia_beta_ablation_r5}
\resizebox{\linewidth}{!}{%
\begin{tabular}{ccc|ccc|ccc}
\toprule
\multicolumn{3}{l|}{$\mathbf{\beta}$\textbf{-term}} &
\multicolumn{3}{c|}{T$\rightarrow$V} &
\multicolumn{3}{c}{V$\rightarrow$T} \\
\cmidrule(lr){4-6}\cmidrule(lr){7-9}
\small{$\beta_r$} & \small{$\beta_c$} & \small{$\beta_g$} & \small{R@1 $\uparrow$} & \small{R@5 $\uparrow$} & \small{MnR $\downarrow$} & \small{R@1 $\uparrow$} & \small{R@5 $\uparrow$} & \small{MnR $\downarrow$} \\
\midrule
1&1&1 & \textbf{48.0} & \textbf{74.1} & 14.2 & 46.9 & 72.7 & 10.6 \\
1&0&0 & 47.6 & 72.7 & 14.3 & 46.9 & \textbf{73.8} & \textbf{9.7} \\
0&1&0 & 47.7 & 73.4 & 13.5 & 47.2 & 73.0 & 9.9 \\
0&0&1 & 47.1 & 72.8 & 14.0 & \textbf{47.5} & 73.0 & 10.4 \\
1&1&0 & 47.3 & 72.9 & 13.6 & 45.8 & 72.9 & 9.9 \\
1&0&1 & 47.0 & 72.9 & 13.8 & 46.7 & 71.9 & 10.8 \\
0&1&1 & 46.5 & 73.1 & \textbf{13.4} & 46.7 & 72.8 & 10.0 \\
\bottomrule
\end{tabular}%
}
\end{table}

\begin{table*}[t]
\centering
\fontsize{9}{11}\selectfont
\setlength{\tabcolsep}{3.9pt}
\renewcommand{\arraystretch}{1.12}
\caption{Unified comparison of model scale/compute and retrieval performance on MSR-VTT (1k-A) and DiDeMo.
FLOPs are computed with fvcore (FlopCountAnalysis) using a profiling batch that follows each repository’s retrieval configuration.
GFLOPs/sample is computed by dividing the total FLOPs of one forward pass (for a profiling batch) by the batch size.
Frames indicate the number of frames used in profiling (All-in-One=3, ALPRO=8, CLIP4Clip/TAME=12). ``--'' indicates metrics not reported.} \label{tab:unified_params_flops_with_frames_ordered}
\begin{tabular}{lccc|cccc|cccc}
\toprule
Model &
Frames & Params (M) & GFLOPs/sample &
\multicolumn{4}{c|}{MSR-VTT (1k-A)} &
\multicolumn{4}{c}{DiDeMo} \\
\cmidrule(lr){5-8}\cmidrule(lr){9-12}
& & & &
R@1 $\uparrow$ & R@5 $\uparrow$ & R@10 $\uparrow$ & MnR $\downarrow$ &
R@1 $\uparrow$ & R@5 $\uparrow$ & R@10 $\uparrow$ & MnR $\downarrow$ \\
\midrule
All-in-One~\cite{allinone} & 3  & 109.87 & 63.91  & 37.9 & 68.1 & 77.1 & - & 32.7 & 61.4 & 73.5 & -   \\
ALPRO~\cite{alpro}      & 8  & 231.48 & 208.36 & 33.9 & 60.7 & 73.2 & - & 35.9 & 67.5 & 78.8 & -   \\
CLIP4Clip  & 12 & 163.93 & 54.12  & 44.0 & 72.0 & \textbf{81.6} & 16.0   & 42.8 & 68.5 & 79.2 & 18.9 \\
TAME & 12 & 205.40 & 77.59  & \textbf{48.0} & \textbf{74.1} & \textbf{81.6} & \textbf{14.2}   & \textbf{45.5} & \textbf{72.9} & \textbf{80.8} & \textbf{14.8} \\
\bottomrule
\end{tabular}
\end{table*}

\begin{table*}[t]
\centering
\fontsize{8}{10}\selectfont   % 또는 \footnotesize
\setlength{\tabcolsep}{3.8pt} % (5.5 -> 3.8)
\renewcommand{\arraystretch}{1.05} % (1.15 -> 1.05)
\caption{Frame permutation ablation on MSR-VTT (1k-A): Text$\rightarrow$Video (T$\rightarrow$V) raw performance.}
\label{tab:perm_ablation_t2v_raw}
\begin{tabular}{l|cccc|ccccc|ccccc}
\toprule
\textbf{Model} & 
\multicolumn{4}{c|}{\textbf{Ordinary}} &
\multicolumn{5}{c|}{\textbf{Reverse}} &  
\multicolumn{5}{c}{\textbf{Random}} \\
\cmidrule(lr){2-5} \cmidrule(lr){6-10} \cmidrule(lr){11-15}
\multicolumn{1}{c|}{} 
& R@1 $\uparrow$ & R@5 $\uparrow$ & R@10 $\uparrow$ & MnR $\downarrow$ 
& R@1 $\uparrow$ & R@5 $\uparrow$ & R@10 $\uparrow$ & MnR $\downarrow$ & $\Delta$R@1
& R@1 $\uparrow$ & R@5 $\uparrow$ & R@10 $\uparrow$ & MnR $\downarrow$ & $\Delta$R@1 \\
\midrule
CLIP4Clip &
44.0 & 72.0 & 81.6 & 16.0 &
43.6 & 71.9 & \textbf{82.0} & 16.1 & -0.9\% &
43.6 & 72.0 & \textbf{82.0} & 16.1 & -0.9\% \\
TAME &
\textbf{48.0} & \textbf{74.1} & 81.6 & \textbf{14.2} &
\textbf{47.1} & \textbf{73.9} & 81.6 & \textbf{14.2} & -1.9\%  &
\textbf{47.6} & \textbf{74.0} & 81.7 & \textbf{14.3} & -0.8\% \\
\bottomrule
\end{tabular}
\end{table*}

% -------------------------------
% Masked-frame (Black-out) test
% -------------------------------
\begin{table*}[t]
\centering
\small
\setlength{\tabcolsep}{4.6pt}
\renewcommand{\arraystretch}{1.12}
\caption{Masked-frame (black-out) test on MSR-VTT (1k-A): Text$\rightarrow$Video (T$\rightarrow$V) average raw performance under different mask ratios.}
\label{tab:blackout_mask_t2v_avg}
\begin{tabular}{c|ccccc|ccccc}
\toprule
\textbf{Mask ratio} &
\multicolumn{5}{c|}{\textbf{CLIP4Clip}} &
\multicolumn{5}{c}{\textbf{TAME}} \\
\cmidrule(lr){2-6}\cmidrule(lr){7-11}
& R@1 $\uparrow$ & R@5 $\uparrow$ & R@10 $\uparrow$ & MnR $\downarrow$ & $\Delta$R@1 (vs.\ 0.0)
& R@1 $\uparrow$ & R@5 $\uparrow$ & R@10 $\uparrow$ & MnR $\downarrow$ & $\Delta$R@1 (vs.\ 0.0) \\
\midrule
0.0 & 44.00 & 72.00 & 81.60 & 16.00 & --      & 48.00 & 74.10 & 81.60 & 14.20 & -- \\
0.1 & 43.20 & 70.96 & 81.30 & 16.06 & -1.82\% & 46.48 & 73.52 & 81.54 & 14.28 & -3.17\% \\
0.2 & 42.34 & 70.40 & 80.54 & 16.34 & -3.77\% & 46.10 & 72.60 & 81.20 & 14.40 & -3.96\% \\
0.3 & 42.02 & 69.36 & 79.30 & 17.04 & -4.50\% & 45.52 & 71.44 & 81.24 & 14.88 & -5.17\% \\
0.4 & 40.10 & 68.08 & 77.82 & 18.30 & -8.86\% & 44.10 & 70.36 & 80.28 & 15.50 & -8.13\% \\
0.5 & 38.84 & 66.60 & 77.16 & 18.92 & -11.73\% & 43.38 & 70.80 & 80.08 & 16.08 & -9.63\% \\
0.6 & 35.68 & 62.00 & 72.88 & 24.16 & -18.91\% & 40.18 & 68.14 & 78.12 & 19.48 & -16.29\% \\
\bottomrule
\end{tabular}
\end{table*}

\subsection{Ablation on MoE Layers and Experts}

We evaluate different placements of the MoE layers in both the text and vision encoders, as shown in Table~\ref{tab:ablation_layer_placement_compact}. Concentrating MoE at the top two backbone layers \(\{11,12\}\) provides the most balanced performance in terms of R@1 and R@5 (R@1=48.0, R@5=74.1). Using only the top layer \(\{12\}\) yields a close R@1 of 47.9 but a slightly lower R@10. By contrast, distributing MoE more broadly (e.g., \(\{7\text{--}12\}\), \(\{8,10,12\}\)) does not improve recall and tends to increase variance in mean rank (MnR). These results suggest that lower layers mainly encode low-level patterns, whereas upper layers provide more disentangled semantic information where expert modules can specialize most effectively. Therefore, rather than distributing MoE throughout the backbone, concentrating it near the upper layers is a more effective design choice for leveraging expert capacity at the representation level.

Table~\ref{tab:ablation_joint_equal_experts} presents the ablation on the number of experts. We observe that using four experts yields the most balanced performance overall, while changing the count to 2, 6, or 8 does not lead to consistent improvements. In all experiments, we set the number of experts in the visual and text encoders to be identical, and the setting \(N_E = 4\) is used as the default configuration.
% Table~\ref{tab:ablation_joint_equal_experts} presents the ablation on the number of experts. The results show that using four experts provides the most balanced performance overall, while changing the count to 2, 6, or 8 yields no consistent gains. This suggests that \(E_v=E_t=4\) is a practical setting.

\subsection{Ablation on Frame--Temporal Tokens}
Table~\ref{tab:ablation_ft_count} shows that using two FT tokens yields the strongest R@1/R@5 (48.0/74.1 with MdR=2.0) and the most balanced overall profile, while one token performs slightly lower (notably higher MnR) and three tokens raise R@10 (82.1) without consistent gains in R@1/R@5, suggesting FT=2 as a practical default.

\subsection{Ablation on CTIA \texorpdfstring{$\beta$}{beta}-Terms}
Table~\ref{tab:ctia_beta_ablation_r5} ablates the CTIA fusion terms by enabling each relevance component ($\mathbf{r}$, $\mathbf{r}_{\mathrm{conv}}$, $\mathbf{r}_{\mathrm{graph}}$) with binary coefficients $(\beta_r,\beta_c,\beta_g)$.
Using all terms jointly $(1,1,1)$ achieves the best Text$\rightarrow$Video performance (48.0/74.1 in R@1/R@5), indicating that raw evidence, local smoothing, and graph propagation provide complementary cues.
Single-term settings remain competitive but show different tendencies: $\mathbf{r}_{\mathrm{conv}}$ yields the best MnR on Text$\rightarrow$Video among single terms (13.5), while $\mathbf{r}_{\mathrm{graph}}$ gives the best Video$\rightarrow$Text R@1 (47.5).
Overall, $(1,1,1)$ provides the most balanced bidirectional profile and is used as the default.

\subsection{Unified Model-Scale and Compute Comparison}
Table~\ref{tab:unified_params_flops_with_frames_ordered} provides a unified comparison between representative unified video--language models (All-in-One~\cite{allinone}, ALPRO~\cite{alpro}) and CLIP-based dual-encoder approaches (CLIP4Clip, TAME) in terms of model scale (Params), compute (GFLOPs/sample), and retrieval performance on MSR-VTT (1k-A) and DiDeMo. FLOPs are computed with fvcore (FlopCountAnalysis) under each repository's retrieval configuration, and the profiling frame counts follow common practice (All-in-One=3, ALPRO=8, CLIP4Clip/TAME=12). Importantly, while our MoE design increases the \emph{total} number of trainable parameters by introducing multiple experts, sparse Top-$K$ routing activates only a small subset of experts per token/sample; therefore, the per-sample compute does not scale linearly with the total number of experts $E$, and is instead primarily governed by $K$ and the number of MoE-inserted blocks (as well as any additional token/mixing modules). Under this unified setting, the unified baselines exhibit a markedly higher compute cost (e.g., ALPRO: 208.36 GFLOPs/sample) yet lower reported R@K than CLIP4Clip and TAME. In contrast, TAME achieves consistent gains over CLIP4Clip with a moderate increase in scale/compute (163.93M/54.12 $\rightarrow$ 205.40M/77.59), improving MSR-VTT R@1 from 44.0 to 48.0 and reducing MnR from 16.0 to 14.2, while also improving DiDeMo R@1 from 42.8 to 45.5 and MnR from 18.9 to 14.8. Moreover, TAME outperforms the larger unified baseline ALPRO while requiring $\sim$2.7$\times$ fewer GFLOPs per sample (208.36 $\rightarrow$ 77.59), demonstrating a favorable accuracy--compute trade-off.

\subsection{Frame Permutation Ablation (Temporal-Order Robustness)}
Table~\ref{tab:perm_ablation_t2v_raw} reports a frame permutation ablation on MSR-VTT (1k-A) under the raw T$\rightarrow$V protocol. We fix the sampled frame set for each video and permute only the temporal order at inference (ordinary/reverse/random), thereby isolating the effect of frame ordering from changes in visual evidence. Overall, both methods show limited sensitivity to order perturbations; however, TAME exhibits a slightly larger degradation under the fully reversed order. Specifically, CLIP4Clip decreases by 0.9\% in R@1 (44.0$\rightarrow$43.6), whereas TAME decreases by 1.9\% (48.0$\rightarrow$47.1). In contrast, under random permutation, the R@1 change of TAME is small (-0.8\%, 48.0$\rightarrow$47.6) and comparable to the baseline (-0.9\%). These results indicate that performance degradation remains limited under temporal-order perturbations, and the overall effect of our method is largely preserved.

This trend is consistent with the design of TAME: we perform content-based cross-frame interaction and aggregation (global mixing over FT tokens and CTIA), while still being able to exploit directional temporal cues to some extent, which can explain why a complete reversal yields a slightly larger drop than a random shuffle. Moreover, under the widely used sparse-frame MSR-VTT protocol, many queries emphasize overall video semantics rather than strict event chronology; therefore, this ablation is best interpreted as a robustness-to-order test (with mild order sensitivity).

\subsection{Masked-Frame Ablation (Robustness to Missing Temporal Evidence)}
Table~\ref{tab:blackout_mask_t2v_avg} reports a masked-frame (black-out) ablation on MSR-VTT (1k-A) under the raw T$\rightarrow$V protocol. We keep the sampled frame set and the retrieval pipeline unchanged, and apply corruption only at inference by randomly selecting a portion of the sampled frames and replacing them with a black-out mask (mask ratio $\in\{0.1,0.2,0.3,0.4,0.5,0.6\}$). This test evaluates robustness to missing/occluded temporal evidence, rather than temporal-order sensitivity.

As the mask ratio increases, both methods degrade (lower R@K and higher MnR), with a noticeably larger drop beyond 0.4. Importantly, TAME maintains higher absolute performance than CLIP4Clip at all mask ratios. Moreover, under heavy masking, TAME shows a smaller relative R@1 drop compared to the unmasked setting: at 0.5, CLIP4Clip decreases by 11.73\% (44.0$\rightarrow$38.84), whereas TAME decreases by 9.63\% (48.0$\rightarrow$43.38); at 0.6, CLIP4Clip decreases by 18.91\% (44.0$\rightarrow$35.68) versus 16.29\% for TAME (48.0$\rightarrow$40.18). Ranking quality is also better preserved by TAME under severe corruption, as reflected by a smaller MnR increase at 0.6 (+5.24 for CLIP4Clip vs.\ +3.40 for TAME).

These results suggest that TAME is less sensitive when a substantial portion of frame-level evidence is removed. Our final similarity integrates a global video--text baseline score with a framewise evidence aggregation score, which reduces over-reliance on any single frame cue and enables more stable consolidation of the remaining informative frames under partial evidence removal. In addition, blacked-out frames typically yield weak similarity signals due to the loss of visual content, and thus receive low influence during framewise consolidation, helping maintain ranking stability when temporal evidence is partially missing.

\section{Conclusion}
In this paper, we presented TAME, a temporal-aware extension of CLIP for text--video retrieval that bridges image-level pre-training and video-level semantics. Building on a CLIP backbone, TAME couples frame-wise Mixture-of-Experts (MoE) layers with frame-consistent routing, Frame--Temporal (FT) tokens, and a cross-temporal interaction and aggregation mechanism (CTIA) to model temporal structure without sacrificing stable vision--language alignment.

Extensive experiments on TVR benchmarks demonstrate that TAME consistently outperforms CLIP-based baselines, with notable gains on MSR-VTT and additional improvements on DiDeMo, MSVD, LSMDC and ActivityNet. These results suggest that the consistent performance improvements observed across diverse benchmarks with different characteristics indicate our method is not a dataset-specific technique, but rather a generalizable approach.

\noindent\textbf{Limitations and Future Work.}
TAME improves retrieval by expanding encoder capacity via sparse MoE and organizing frame evidence via lightweight temporal tokens and CTIA, but it introduces trade-offs in routing stability and temporal-token overhead.
Our ablations suggest that MoE benefits require careful stabilization (e.g., load balancing), and temporal mixing must be applied selectively to control memory cost.
Moreover, CTIA is efficient and interpretable, yet it relies on the quality of framewise similarities and a small set of hyperparameters.
Future work will focus on more efficient temporal mixing for long videos, more robust/adaptive routing, and dynamic compute allocation (e.g., learned frame sampling or token budgeting), as well as extending CTIA-style aggregation to temporally grounded or streaming retrieval. In addition, we will explore efficiency-oriented model compression strategies such as pruning, quantization, and knowledge distillation to develop lightweight variants of TAME that retain high retrieval accuracy with reduced model size and inference cost.

\section*{Acknowledgements}
This work was partly supported by the Institute of Information and Communications Technology Planning and Evaluation (IITP) grant funded by the Korea government (MSIT) under Grant No. 2710017875 (Development of multimodal data input-based search augmentation generation technology, 50\%) and Grant No. IITP-2025-RS-2023-00254529 (Metaverse support program to nurture the best talents, 25\%), and by the National Research Foundation of Korea (NRF) grant funded by the Korea government (MSIT) under Grant No. RS-2025-00553785 (Research on the core technology of data integration and augmentation for OTT user analysis, 25\%).

{
    \small
    \IfFileExists{refs.bib}{%
        \bibliographystyle{ieeetr}
        \bibliography{refs}
    }{%
        \section*{References}
        Reference entries are omitted in this one-file preview because
        \texttt{refs.bib} was not uploaded.
    }%
}

\end{document}